\documentclass[10pt,twocolumn,letterpaper]{article}

\usepackage[pagenumbers]{cvpr} 

\usepackage{mmstyle}
\usepackage{graphicx}
\usepackage{amsmath}
\usepackage{amssymb}
\usepackage{booktabs}
\usepackage{multirow}
\usepackage{mathrsfs}
\usepackage[table]{xcolor}
\usepackage{marvosym}
\usepackage[hang,flushmargin]{footmisc}

\newcommand{\rankone}{add8e6}
\newcommand{\ranktwo}{b5dce9}
\newcommand{\rankthree}{bde0eb}
\newcommand{\rankfour}{c6e4ee}
\newcommand{\rankfive}{cee8f0}
\newcommand{\ranksix}{d6ecf3}
\newcommand{\rankseven}{deeff5}
\newcommand{\rankeight}{e6f3f8}
\newcommand{\ranknine}{eff7fa}
\newcommand{\rankten}{f7fbfd}

\usepackage[accsupp]{axessibility} 

\usepackage[pagebackref,breaklinks,colorlinks]{hyperref}

\usepackage[capitalize]{cleveref}
\crefname{section}{Sec.}{Secs.}
\Crefname{section}{Section}{Sections}
\Crefname{table}{Table}{Tables}
\crefname{table}{Tab.}{Tabs.}

\def\cvprPaperID{2642} 
\def\confName{CVPR}
\def\confYear{2023}

\begin{document}

\title{OmniObject3D: Large-Vocabulary 3D Object Dataset for \\ Realistic  Perception, Reconstruction and Generation }

\author{Tong Wu$^{2}$, 
Jiarui Zhang$^{1,3}$,
Xiao Fu$^1$,
Yuxin Wang$^{1,4}$,
Jiawei Ren$^5$, 
Liang Pan$^5$, \\
Wayne Wu$^1$, 
Lei Yang$^{1,3}$, 
Jiaqi Wang$^1$,
Chen Qian$^1$, 
Dahua Lin$^{1,2}{\textsuperscript{\Letter}}$, 
Ziwei Liu$^{5}{\textsuperscript{\Letter}}$\\
$^1$Shanghai Artificial Intelligence Laboratory,
$^2$The Chinese University of Hong Kong,
$^3$SenseTime Research, \\
$^4$Hong Kong University of Science and Technology,
$^5$S-Lab, Nanyang Technological University\\ 
\tt\small
\{wt020,dhlin\}@ie.cuhk.edu.hk,
zjr954@163.com,
ywangom@connect.ust.hk,\\
\tt\small
wuwenyan0503@gmail.com, 
\{fuxiao,wangjiaqi,qianchen\}@pjlab.org.cn, \\
\tt\small
yanglei@sensetime.com,
jiawei011@e.ntu.edu.sg,
\{liang.pan,ziwei.liu\}@ntu.edu.sg \\
}


\twocolumn[{
\renewcommand\twocolumn[1][]{#1}
\maketitle
\begin{center}
    \centering
    \vspace{-20pt}
    \includegraphics[width=0.95\linewidth]{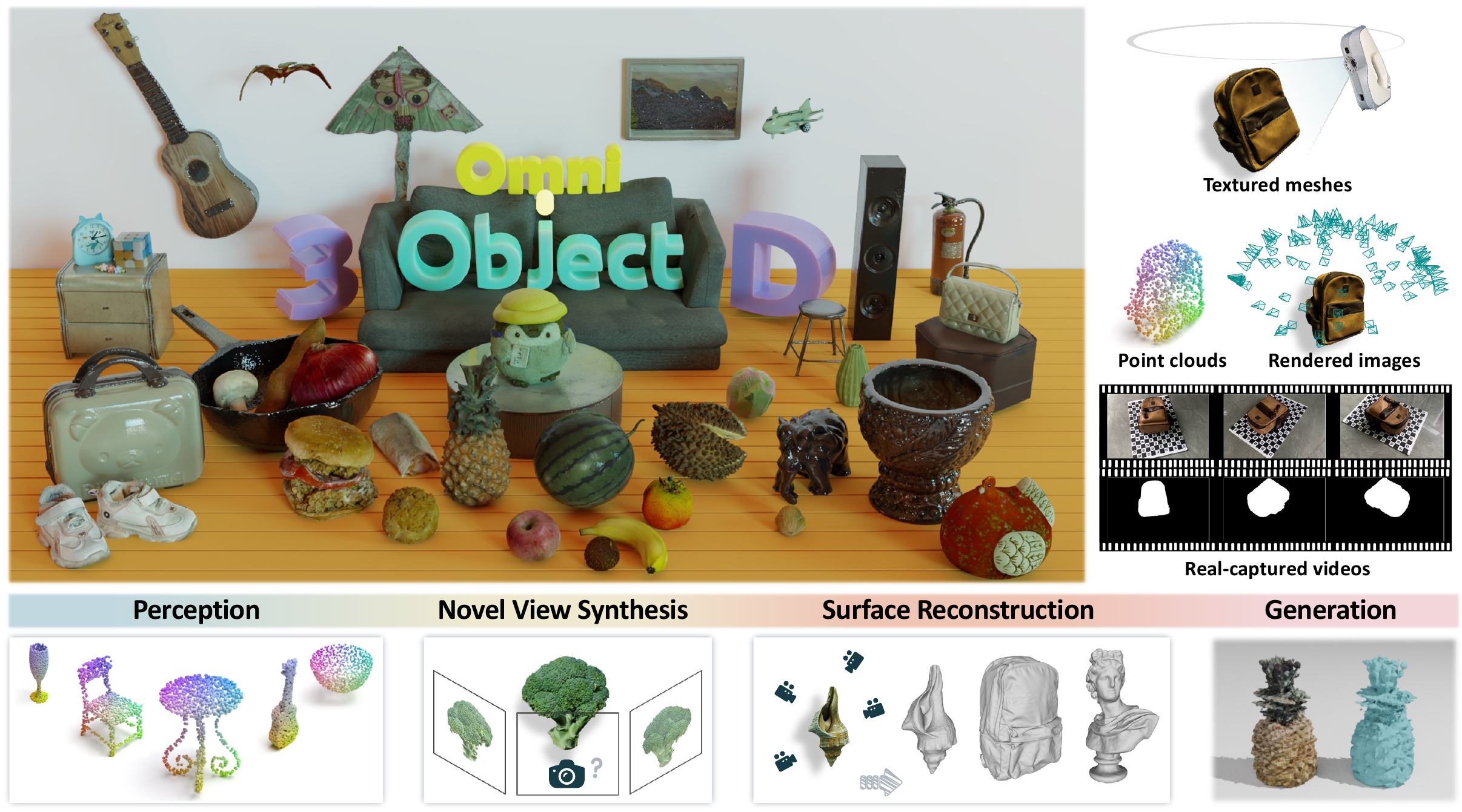}
    \setlength{\abovecaptionskip}{0mm}
    \captionof{figure}{\small
	\textbf{OmniObject3D is a large vocabulary 3D object dataset with massive high-quality real-scanned 3D objects and rich annotations.} It  supports various research topics, \eg, perception, novel view synthesis, neural surface reconstruction, and 3D generation.  
	}
	\label{fig:teaser}
\end{center}
}]

{\let\thefootnote\relax\footnotetext{\noindent\textsuperscript{\Letter}Corresponding authors. \url{https://omniobject3d.github.io/}}}

\begin{abstract}
   Recent advances in modeling 3D objects mostly rely on synthetic datasets due to the lack of large-scale real-scanned 3D databases. 
   To facilitate the development of 3D perception, reconstruction, and generation in the real world, we propose \textbf{OmniObject3D}, a large vocabulary 3D object dataset with massive high-quality real-scanned 3D objects. 
   OmniObject3D has several appealing properties: 
   \textbf{1) Large Vocabulary:} It comprises 6,000 scanned objects in 190 daily categories, sharing common classes with popular 2D datasets (e.g., ImageNet and LVIS), benefiting the pursuit of generalizable 3D representations.   
   \textbf{2) Rich Annotations:} Each 3D object is captured with both 2D and 3D sensors, providing textured meshes, point clouds, multi-view rendered images, and multiple real-captured videos.
   \textbf{3) Realistic Scans:} The professional scanners support high-quality object scans with precise shapes and realistic appearances.
   With the vast exploration space offered by OmniObject3D, we carefully set up four evaluation tracks: \textbf{a)} robust 3D perception, \textbf{b)} novel-view synthesis, \textbf{c)} neural surface reconstruction, and \textbf{d)} 3D object generation.
   Extensive studies are performed on these four benchmarks, revealing new observations, challenges, and opportunities for future research in realistic 3D vision.
\end{abstract}

\vspace{-15pt}
\section{Introduction}

Sensing, understanding, and synthesizing realistic 3D objects is a long-standing problem in computer vision, with rapid progress emerging in recent years. However, a majority of the technical approaches rely on unrealistic synthetic datasets~\cite{chang2015shapenet,wu20153d,fu20213d} due to the absence of a large-scale real-world 3D object database. However, the appearance and distribution gaps between synthetic and real data cannot be compensated for trivially, hindering their real-life applications. Therefore, it is imperative to equip the community with a large-scale and high-quality 3D object dataset from the real world, which can facilitate a variety of 3D vision tasks and downstream applications.

Recent advances partially fulfill the requirements while still being unsatisfactory. As shown in Table~\ref{tab:dataset_comparison}, CO3D~\cite{reizenstein2021co3d} contains 19k videos capturing objects from 50 MS-COCO categories, while only 20\% of the videos are annotated with accurate point clouds reconstructed by COLMAP~\cite{schonberger2016sfm}. Moreover, they do not provide textured meshes. GSO~\cite{downs2022google} has 1k scanned objects while covering only 17 household classes. AKB-48~\cite{liu2022akb48} focuses on robotics manipulation with 2k articulated object scans in 48 categories, but the focus on articulation leads to a relatively narrow semantic distribution, failing to support general 3D object research.

To boost the research on general 3D object understanding and modeling, we present \textbf{OmniObject3D}: a large-vocabulary 3D object dataset with massive high-quality, real-scanned 3D objects.
Our dataset has several appealing properties: 
\textbf{1) Large Vocabulary:} It contains 6,000 high-quality textured meshes scanned from real-world objects, which, to the best of our knowledge, is the largest among real-world 3D object datasets with accurate 3D meshes. It comprises 190 daily categories, sharing common classes with popular 2D and 3D datasets (\eg, ImageNet~\cite{deng2009imagenet}, LVIS~\cite{gupta2019lvis}, and ShapeNet~\cite{chang2015shapenet}), incorporating most daily object realms (See Figure~\ref{fig:teaser} and Figure~\ref{fig:statistics}).
\textbf{2) Rich Annotations:} Each 3D object is captured with both 2D and 3D sensors, providing textured 3D meshes, sampled point clouds, posed multi-view images rendered by Blender~\cite{blender}, and real-captured video frames with foreground masks and COLMAP camera poses. 
\textbf{3) Realistic Scans:} The object scans are of high fidelity thanks to the professional scanners, bearing precise shapes with geometric details and realistic appearance with high-frequency textures.

Taking advantage of the vast exploration space offered by OmniObject3D, we carefully set up four evaluation tracks: \textbf{a)} robust 3D perception, \textbf{b)} novel-view synthesis, \textbf{c)} neural surface reconstruction, and \textbf{d)} 3D object generation. 
Extensive studies are performed on these benchmarks:
First, the high-quality, real-world point clouds in OmniObject3D allow us to perform \textit{robust 3D perception} analysis on both out-of-distribution (OOD) styles and corruptions, two major challenges in point cloud OOD generalization.  
Furthermore, we provide massive 3D models with multi-view images and precise 3D meshes for \textit{novel-view synthesis} and \textit{neural surface reconstruction}. The broad diversity in shapes and textures offers a comprehensive training and evaluation source for both scene-specific and generalizable algorithms.
Finally, we equip the community with a database for large vocabulary and realistic \textit{3D object generation}, which pushes the boundary of existing state-of-the-art generation methods to real-world 3D objects.
The four benchmarks reveal new observations, challenges, and opportunities for future research in realistic 3D vision. 

\section{Related Works}
\label{sec:related}


\begin{table}[t] 
  \centering
  \small
  \setlength{\abovecaptionskip}{0mm}
  \caption{\textbf{A comparison between OmniObject3D and other commonly-used 3D object datasets.}
  R$^{\text{lvis}}$ denotes the ratio of the 1.2k LVIS~\cite{gupta2019lvis} categories being covered.}
    \setlength\tabcolsep{3.2pt}
    \resizebox{.49\textwidth}{!}{
    \begin{tabular}{l||cccccc}
    \toprule
    Dataset  & Real  & Full Mesh & Video & \# Objs & \# Cats & R$^{\text{lvis}}$ (\%) \\
    \midrule
    ShapeNet~\cite{chang2015shapenet}  &       & \checkmark &       & 51k   & 55    & 4.1 \\
    ModelNet~\cite{wu20153d} &       & \checkmark &       & 12k   & 40    & 2.4 \\
    3D-Future~\cite{fu20213d} &       & \checkmark &       & 16k   & 34    & 1.3 \\
    ABO~\cite{collins2022abo}   &       & \checkmark &       & 8k    & 63    & 3.5 \\
    Toys4K~\cite{stojanov2021toys4k} &       & \checkmark &       & 4k    & 105   & 7.7 \\
    CO3D V1 / V2~\cite{reizenstein2021co3d}  & \checkmark &       & \checkmark & 19 / 40k   & 50   & 4.2 \\
    DTU~\cite{aanaes2016large}   & \checkmark & \checkmark &       & 124   & NA    & 0 \\
    ScanObjectNN~\cite{uy2019revisiting}  & \checkmark &     &       & 15k   & 15    & 1.3 \\
    GSO~\cite{downs2022google}   & \checkmark & \checkmark &       & 1k    & 17    & 0.9 \\
    AKB-48~\cite{liu2022akb48} & \checkmark & \checkmark &       & 2k    & 48    & 1.8 \\
    \textbf{Ours} & \checkmark & \checkmark & \checkmark & 6k    & 190   & 10.8 \\
    \bottomrule
    \end{tabular}}
  \label{tab:dataset_comparison}%
  \vspace{-15pt}
\end{table}%

\begin{figure*}[t]
    \centering
    \includegraphics[width=0.95\linewidth]{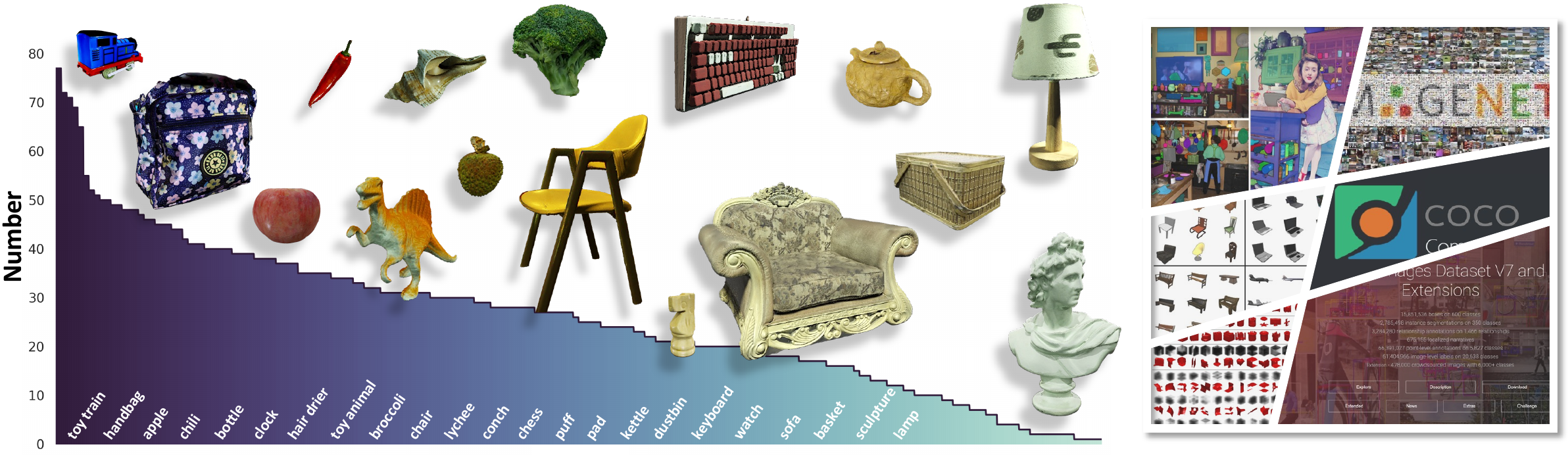}
    \setlength{\abovecaptionskip}{-0.5mm}
    \caption{\small
    \textbf{Semantic distribution of the OmniObject3D dataset.} It covers 190 daily categories with a long-tailed distribution, sharing common classes with popular 2D and 3D datasets.
    }
    \label{fig:statistics}
    \vspace{-15pt}
\end{figure*}

\noindent{\textbf{3D Object Datasets.}}
The acquisition of a large-scale realistic 3D database is usually expensive and challenging. Many widely-used 3D datasets prefer to collect synthetic CAD models from online repositories~\cite{chang2015shapenet,wu20153d,stojanov2021toys4k}, for example, ShapeNet~\cite{chang2015shapenet} has 51,300 3D CAD models in 55 categories, and ModelNet40~\cite{wu20153d} consists of 12,311 models in 40 categories.
Recent works, \eg, 3D-FUTURE~\cite{fu20213d} and ABO~\cite{collins2022abo}, introduce high-quality CAD models with rich geometric details and informative textures.
However, due to the inevitable gap between synthetic and real objects, the community is still eager for a large-scale 3D object dataset in the real world.
DTU~\cite{aanaes2016large} and BlendedMVS~\cite{yao2020blendedmvs} are photo-realistic datasets designed for multi-view stereo benchmarks, while they are small in scale and lack category annotations.
ScanObjectNN~\cite{uy2019revisiting} is a real-world point cloud object dataset based on scanned indoor scenes, containing around 15,000 objects with colored point clouds in 15 categories. However, the point clouds are incomplete and noisy, and multiple objects usually co-exist in one scene.
GSO~\cite{downs2022google} has 1,030 scanned objects with fine geometries and textures in 17 household items, and AKB-48~\cite{liu2022akb48} focuses on robotics manipulation with 2,037 articulated object models in 48 articulated object categories. However, the relatively narrow semantic scope of GSO and AKB-48 hinders their applications for more general 3D research. CO3D v1~\cite{reizenstein2021co3d} contains 19,000 object-centric videos, while only 20\% of them are annotated with accurate point clouds reconstructed by COLMAP~\cite{schonberger2016sfm}, and they do not provide meshes or textures. 
In contrast, the proposed OmniObject3D dataset comprises 6,000 3D objects scanned by professional devices with meshes, textures, and multi-view photos in 190 categories, fulfilling the requirements of a wide range of research objectives.
A detailed comparison is presented in Table~\ref{tab:dataset_comparison}.

\noindent{\textbf{Robust 3D Perception.}}
Robustness to out-of-distribution (OOD) data is important in point cloud perception. Two main challenges include OOD styles (\eg, differences between CAD models and real-world objects) and OOD corruptions (\eg, random point jittering or missing due to sensory inaccuracy). A branch of works~\cite{qi2016pointnet, wang2019dgcnn, chen2020pointmixup, kim2021pointwolf} studies the OOD corruptions and proposes standard corruption test suites~\cite{taghanaki2020robustpointset, ren2022modelnet-c}, while they fail to take account of OOD styles. Another branch of works~\cite{reizenstein2021co3d,ahmadyan2021objectron} evaluates the sim-to-real domain gap by training models on clean synthetic datasets~\cite{wu20153d} and testing them on noisy real-world test sets~\cite{uy2019revisiting}, while OOD styles and corruptions cannot be disentangled under this setting for an independent analysis. In this work, we leverage high-quality, real-world point clouds from OmniObject3D to systematically measure the robustness against the OOD style and OOD corruptions, providing the first benchmark for fine-grained evaluation of the point cloud perception robustness. 

\noindent{\textbf{Neural Radiance Field and Neural Surface Reconstruction.}}
Neural radiance field (NeRF)~\cite{mildenhall2020nerf} represents a scene with a fully-connected deep network (MLPs), which takes in hundreds of sampled points along each camera ray and outputs the predicted color and density. We can synthesize the image of an unseen view from a trained model via volume rendering.
Inspired by the success of NeRF, massive follow-up efforts have been made to improve its quality~\cite{barron2021mip, DorVerbin2022RefNeRFSV, BenMildenhall2021NeRFIT, JonathanTBarron2021MipNeRF3U} and efficiency~\cite{yu2021plenoxels, sun2021direct, chen2022tensorf, mueller2022instant}.
A branch of works~\cite{AlexYu2021pixelNeRFNR,AnpeiChen2021MVSNeRFFG,QianqianWang2021IBRNetLM,YuanLiu2021NeuralRF,reizenstein2021co3d,henzler2021unsupervised} has also explored the generalization ability of NeRF-based frameworks, where they aim to learn priors from deep image features across multiple scenes.
Beyond novel view synthesis, another trend of approaches~\cite{oechsle2021unisurf,yariv2021volume,wang2021neus,francois2021warping,wu2022voxurf} proposes to combine neural radiance field with implicit surface representations like Signed Distance Function (SDF), and they achieve accurate and mask-free surface reconstruction from multi-view images. 
Since dense camera views of scenes are sometimes unavailable, recent advances explore surface reconstruction from sparse views. They address the problem by exploiting generalizable priors cross scenes for a generic surface prediction~\cite{XiaoxiaoLong2022SparseNeuSFG} or taking advantage of the estimated geometry cues estimated by pre-trained networks~\cite{ZehaoYu2022MonoSDFEM}.
OmniObject3D can serve as a large-scale benchmark with realistic photos and accurate meshes for both training and evaluation. The high diversity in shape and appearance offers an opportunity for pursuing more generalizable and robust novel view synthesis and surface reconstruction methods. 

\noindent{\textbf{3D Object Generation.}}
Early approaches~\cite{JiajunWu2016LearningAP,MatheusGadelha20163DSI,PhilippHenzler2018EscapingPC,SebastianLunz2020InverseGG,EdwardJSmith2017ImprovedAS} extend 2D generation frameworks to 3D voxels with a high computational cost. Some others adopt different 3D data formulations, \eg, point cloud~\cite{PanosAchlioptas2017LearningRA,GuandaoYang2019PointFlow3P,LinqiZhou20213DSG,KaichunMo2019StructureNetHG}, octree~\cite{MoritzIbing2022OctreeTA}, and implicit representations~\cite{LarsMescheder2018OccupancyNL,ZhiqinChen2018LearningIF}. However, it is non-trivial to generate complex and textured surfaces. Recent advances~\cite{DarioPavllo2021LearningGM,WenzhengChen2019LearningTP,JunGao2022GET3DAG} explore the generation for textured 3D meshes, where GET3D~\cite{gao2022get3d} is a state-of-the-art approach that generates diverse meshes with rich geometry and textures in two branches. It is a promising but challenging task to train generative models on a large vocabulary and realistic dataset. We evaluate GET3D on our dataset and reveal several challenges and future opportunities.

In supplementary materials, we present more detailed discussions on related works for different tracks.
\section{The OmniObject3D Dataset}
In this section, we describe the data collection, processing, and annotation pipeline of OmniObject3D. We also introduce the statistics and distribution of it.

\subsection{Data Collection, Processing, and Annotation}
\noindent{\textbf{Category List Definition.}}
In order to collect a large amount of 3D objects that are both commonly-distributed and highly diverse, we first pre-define a category list according to several popular 2D and 3D datasets~\cite{deng2009imagenet,lin2014microsoft,gupta2019lvis,kuznetsova2020open,shao2019objects365,chang2015shapenet,wu20153d}. We cover most of the categories that lie within the application scope of the scanners and also dynamically expand the list with reasonable new classes that are absent from the current list during collection. 
We end up with 190 widely-spread categories, which ensures a library with rich texture, geometry, and semantic information.

\noindent{\textbf{Object Collection Pipeline.}} 
We then collect a variety of objects from each category and use professional 3D scanners to obtain high-resolution textured meshes. Specifically, we use the Shining 3D scanner~\footnote{https://www.einscan.com/} and Artec Eva 3D scanner~\footnote{https://www.artec3d.cn/} for objects in different scales.
The scanning time varies with the properties of the object: it takes around 15 minutes to scan a small rigid object with a simple geometry (\eg , an apple, a toy), while it takes up to an hour to obtain a qualified 3D scan for non-rigid, complex, or large objects (\eg , a bed, a kite). 
For around $10\%$ of the objects, we conduct common manipulations (\eg, taking a bite, cutting in pieces) to conform the natural instincts of them.
The 3D scans can faithfully retain the real-world scale of each object, but their poses are not strictly aligned. We thus pre-define a canonical pose for each category and manually align the objects within a category. 
We then check the quality of each scan, and around 83\% high-quality ones out of the total collection are finally reserved in the dataset. 

\noindent\textbf{Image Rendering and Point Cloud Sampling.}
To support a variety of research topics like point cloud analysis, neural radiance fields, and 3D generation, we render multi-view images and sample point clouds based on the collected 3D models. 
We use Blender~\cite{blender} to render object-centric and photo-realistic multi-view images, together with accurate camera poses. The images are rendered from 100 random viewpoints sampled on the upper hemisphere at 800 × 800 pixels. We also produce high-resolution mid-level cues like depth and normal for more research use. 
We then uniformly sample multi-resolution point clouds from each 3D model using the Open3D toolbox~\cite{zhou2018open3d}, with $2^n$ ($n\in \{10, 11, 12, 13, 14\}$) points in each point cloud, respectively.
Besides the data existing in the dataset, we also provide a data generation pipeline. One can easily obtain new data with self-defined camera distributions, lighting, and point sampling methods to meet different requirements.

\noindent{\textbf{Video Capturing and Annotation.}}
After scanning each object, we capture its video with an iPhone 12 Pro mobile phone. The object is placed on or beside a calibration board, and each video covers a full $360^{\circ}$ range around it. Square corners on the calibration board can be recognized by the QR Codes beside it, and we then filter out the blurry frames whose recognized corners are less than 8. We uniformly sample 200 frames, and then COLMAP~\cite{schonberger2016sfm}, a well-known SfM pipeline, is applied to annotate the frames with camera poses. Finally, we use the scales of the calibration board in both the SfM coordinate space and the real world to recover the absolute scale of the SfM coordinate system.
We also develop a two-stage matting pipeline based on the U$^2$Net~\cite{Qin_2020_PR} and FBA~\cite{forte2020fbamatting} matting model to generate the foreground masks for all the frames.
Please refer to supplementary materials for more implementation details.

\subsection{Statistics and Distribution}
With 6,000 3D models in 190 categories, OmniObject3D exhibits a long-tailed distribution with an average of around 30 objects in each category, as shown in Figure~\ref{fig:statistics}. 
It shares many common categories with several famous 2D and 3D datasets~\cite{deng2009imagenet,lin2014microsoft,gupta2019lvis,kuznetsova2020open,shao2019objects365,chang2015shapenet,wu20153d}. For example, it covers 85 categories in ImageNet~\cite{deng2009imagenet} and 130 categories in LVIS~\cite{gupta2019lvis}, which leads to the highest R$^{\text{lvis}}$ in Table~\ref{tab:dataset_comparison}. Most of the categories are covered by the Open Images~\cite{kuznetsova2020open} image-level labels. 
It bears a huge diversity in shapes and appearances. The vast semantic and geometrical exploration spaces enable a wide range of research objectives.


\section{Experiments}
\subsection{Robust 3D Perception}
Object-level point cloud classification is one of the most fundamental tasks in point cloud perception. In this section, we show how OmniObject3D boosts robustness analysis of point cloud classification by disentangling the two critical out-of-distribution (OOD) challenges introduced in Sec.~\ref{sec:related}, \ie, OOD styles and OOD corruptions.

\begin{figure}[t!]
\centering
\includegraphics[width=0.9\columnwidth]{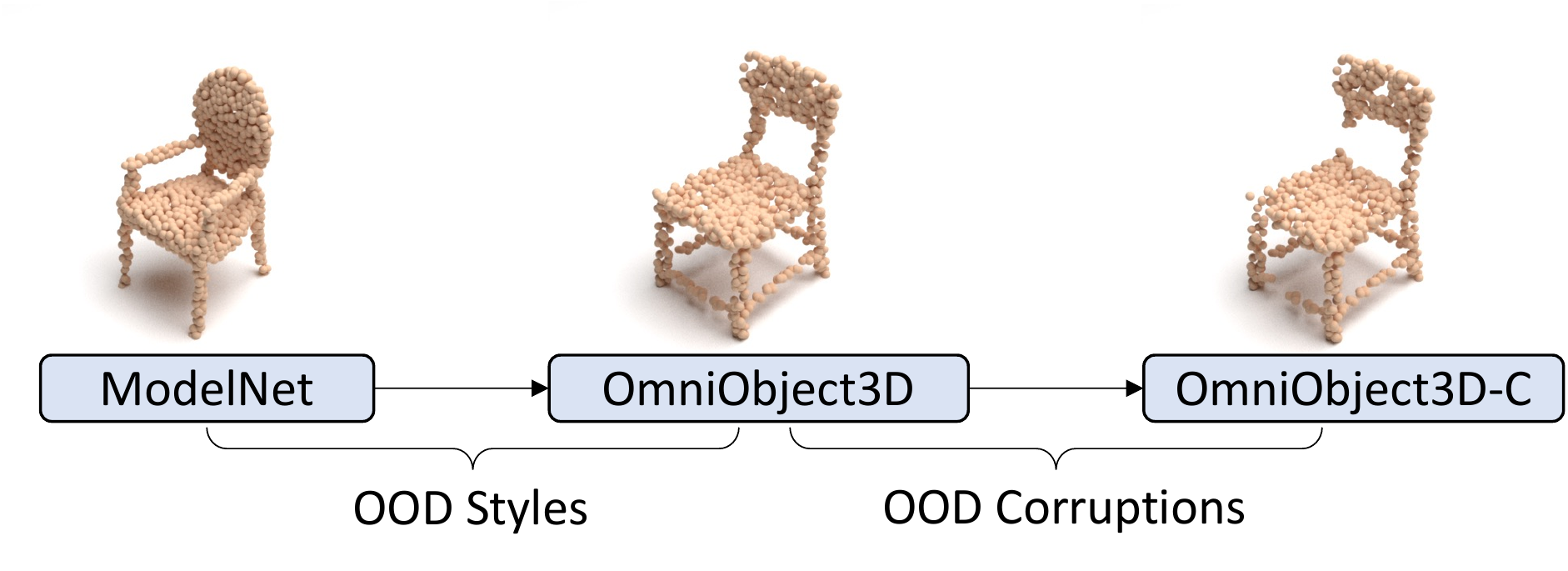}
\setlength{\abovecaptionskip}{0mm}
\caption{OmniObject3D provides the first clean real-world point cloud object dataset and allows fine-grained analysis on robustness to OOD styles and OOD corruptions.``-C": corrupted by common corruptions described in~\cite{ren2022modelnet-c}}
\vspace{-20pt}
\label{fig:robustness_eval}
\end{figure}

Existing robustness evaluation utilizes clean synthetic datasets, \eg, ModelNet~\cite{wu20153d}, for training and sets up two kinds of test sets for evaluation: 

\noindent{\textit{1) Noisy real-world datasets}}, \eg, ScanObjectNN~\cite{uy2019revisiting}, which are cropped from real-world scenes. They are employed to measure the robustness of the sim-to-real domain gap. However, the gap couples both OOD styles and OOD corruptions simultaneously as the cropped point clouds are always noisy, making it impossible to analyze the two robustness challenges independently.

\noindent{\textit{2) Corrupted synthetic datasets}}, \eg, ModelNet-C~\cite{ren2022modelnet-c}, which are artificially perturbed on top of clean synthetic datasets. The evaluation allows for detailed corruption analysis, but they do not reflect the robustness to OOD styles.

\begin{table}
\centering\small
\setlength{\abovecaptionskip}{0mm}
\caption{\textbf{Point cloud perception robustness analysis on OmniObject3D with different architecture designs.} Models are trained on the ModelNet-40 dataset, with $\text{OA}_\text{Clean}$ to be their overall accuracy on the standard ModelNet-40 test set. $\text{OA}_\text{Style}$ on OmniObject3D evaluates the robustness to OOD styles. mCE on the corrupted OmniObject3D-C evaluates the robustness to OOD corruptions. Blue shadings indicate rankings. $\dagger$: results on ModelNet-C~\cite{ren2022modelnet-c}. Full results are presented in the supplementary materials.}
\label{tab:pcd_robustness_full}
\begin{tabular}{lc|ccc}
\toprule
{} &   $\text{mCE}^\dagger\downarrow$  &     $\text{OA}_\text{Clean}\uparrow$ &                               $\text{OA}_\text{Style}\uparrow$ &                              $\text{mCE}\downarrow$ \\
\midrule
DGCNN~\cite{wang2019dgcnn}  & \cellcolor[HTML]{\rankfive}1.000    & 
\cellcolor[HTML]{FFFFFF}0.926 &
\cellcolor[HTML]{FFFFFF}0.448 &              \cellcolor[HTML]{\rankseven}1.000 \\
PointNet~\cite{qi2016pointnet}   & \cellcolor[HTML]{\rankten}1.422 &
\cellcolor[HTML]{FFFFFF}0.907 &
\cellcolor[HTML]{FFFFFF}0.466 &             \cellcolor[HTML]{\rankfive}0.969 \\
PointNet++~\cite{qi2017pointnetplusplus}  & \cellcolor[HTML]{\rankseven}1.072 &
\cellcolor[HTML]{FFFFFF}0.930 &
\cellcolor[HTML]{FFFFFF}0.407 &                         \cellcolor[HTML]{\rankeight}1.066 \\
RSCNN~\cite{liu2019rscnn}      & \cellcolor[HTML]{\ranknine}1.130 & 
\cellcolor[HTML]{FFFFFF}0.923 &
\cellcolor[HTML]{FFFFFF} 0.393 &                           \cellcolor[HTML]{\rankten}1.076 \\
SimpleView~\cite{goyal2021simpleview} & \cellcolor[HTML]{\ranksix}1.047 &
\cellcolor[HTML]{FFFFFF}\textbf{0.939} &
\cellcolor[HTML]{FFFFFF}0.476 &                          \cellcolor[HTML]{\ranksix}0.990 \\
GDANet~\cite{xu2021gdanet}     & \cellcolor[HTML]{\ranktwo}\underline{0.892} &
\cellcolor[HTML]{FFFFFF}0.934 &
\cellcolor[HTML]{FFFFFF}\underline{0.497} &
\cellcolor[HTML]{\rankone} \textbf{0.920} \\
PAConv~\cite{xu2021paconv}     & \cellcolor[HTML]{\rankeight}1.104 &
\cellcolor[HTML]{FFFFFF}0.936 &
\cellcolor[HTML]{FFFFFF} 0.403 &                            \cellcolor[HTML]{\ranknine}1.073 \\
CurveNet~\cite{xiang2021curvenet}   & \cellcolor[HTML]{\rankfour}0.927 &
\cellcolor[HTML]{FFFFFF}\underline{0.938} &
\cellcolor[HTML]{FFFFFF}\textbf{0.500} & 
\cellcolor[HTML]{\ranktwo} \underline{0.929} \\
PCT~\cite{guo2020pct}        & \cellcolor[HTML]{\rankthree}0.925 &
\cellcolor[HTML]{FFFFFF}0.930 &
\cellcolor[HTML]{FFFFFF}0.459 &                          \cellcolor[HTML]{\rankfour}0.940 \\
RPC~\cite{ren2022modelnet-c}        & \cellcolor[HTML]{\rankone}\textbf{0.863} &
\cellcolor[HTML]{FFFFFF}0.930 &
\cellcolor[HTML]{FFFFFF}0.472 &                           \cellcolor[HTML]{\rankthree}0.936 \\
\bottomrule
\end{tabular}
\vspace{-20pt}
\end{table}

None of the existing robustness benchmarks allows for analyzing the robustness to both OOD styles and OOD corruptions in fine granularity. OmniObject3D, on the other hand, as the first clean real-world point cloud object dataset, can help to address the issue. For models trained on ModelNet, we first evaluate their performance on OmniObject3D to examine OOD-style robustness. Then, we create OmniObject3D-C by corrupting OmniObject3D with common corruptions described in~\cite{ren2022modelnet-c} to examine the OOD-corruption robustness. We show a complete robustness evaluation scheme in Figure~\ref{fig:robustness_eval}.
For evaluation metrics, we use the overall accuracy (OA) on OmniObject3D to measure the OOD-style robustness and use DGCNN normalized mCE~\cite{ren2022modelnet-c} to measure the OOD-corruption robustness. 

We benchmark ten state-of-the-art point cloud classification models in Table~\ref{tab:pcd_robustness_full}. We observe that 1) performance on a clean test set has little correlation with OOD-style robustness. For example, SimpleView~\cite{goyal2021simpleview} achieves the best $\text{OA}_\text{Clean}$ but mediocre $\text{OA}_\text{Style}$; 2) Advanced point grouping, \eg, curve-based point grouping in CurveNet~\cite{xiang2021curvenet} and frequency-based point grouping in GDANet~\cite{xu2021gdanet}, are robust not only to OOD corruptions as pointed out in~\cite{ren2022modelnet-c}, but also to OOD styles; 3) OOD style + OOD corruption is a more challenging setting. In particular, RPC, the most robust architecture to OOD corruptions~\cite{ren2022modelnet-c}, shows inferior \text{mCE}. In summary, robust point cloud perception models against both OOD styles and OOD corruptions are still under-explored. Our dataset sheds new light on a comprehensive understanding of point cloud perception robustness. See more results in the supplementary materials.



\subsection{Novel View Synthesis}
\label{sec:nerf}
In this section, we study several representative methods on OmniObject3D for novel view synthesis (NVS) in two settings: 
1) training on a single scene with densely captured images, which is the standard setting for NeRF~\cite{mildenhall2020nerf};
2) learning priors across scenes from our dataset to explore the generalization ability of NeRF-style models.

\begin{table}[t]
  \centering
  \small
  \setlength{\abovecaptionskip}{0mm}
  \caption{\textbf{Single-scene novel view synthesis results.} Three metrics and their standard deviation (SD) across the training set.}
  \setlength\tabcolsep{3.2pt}
    \begin{tabular}{l||ccc}
    \toprule
    Method & PSNR ($\uparrow$) / SD & SSIM ($\uparrow$) / SD & LPIPS ($\downarrow$) / SD \\
    \midrule
    NeRF~\cite{mildenhall2020nerf}  & 34.01 / 3.46 & 0.953 / 0.029 & 0.068 / 0.061 \\
    mip-NeRF~\cite{barron2021mip} & 39.86 / 4.58 & 0.974 / 0.013 & 0.084 / 0.048 \\
    Plenoxels~\cite{yu2021plenoxels} & \textbf{41.04} / 6.84 & \textbf{0.982} / 0.031 & \textbf{0.030} / 0.031 \\
    \bottomrule
    \end{tabular}%
  \label{tab:single_nerf}%
  \vspace{-15pt}
\end{table}%

\begin{table}[t]
  \centering
  \small
  \setlength{\abovecaptionskip}{0mm}
  \caption{\textbf{Cross-scene novel view synthesis results on 10 categories.} `Cat.' and `All*' denote training on each category and training on all categories except the 10 test ones, respectively.}
  \setlength\tabcolsep{3.2pt}
  \resizebox{.49\textwidth}{!} {
    \begin{tabular}{l|c||cccc}
    \toprule
    Method & Train & PSNR ($\uparrow$)  & SSIM ($\uparrow$)  & LPIPS ($\downarrow$) & $\mathcal{L}_{1}^{\text{depth}}$  ($\downarrow$) \\
    \midrule
    \multirow{4}[2]{*}{MVSNeRF~\cite{AnpeiChen2021MVSNeRFFG}} & All* & 17.49 & 0.544 & 0.442 & 0.193 \\
          & Cat. & 17.54 & 0.542 & 0.448 & 0.230 \\
          & All*-ft. & 25.70 & 0.754 & 0.251 & 0.081 \\
          & Cat.-ft. & 25.52 & 0.750 & 0.264 & \textbf{0.076} \\
    \midrule
    \multirow{4}[2]{*}{IBRNet~\cite{QianqianWang2021IBRNetLM}} & All* & 19.39 & 0.569 & 0.399 & 0.423 \\
          & Cat. & 19.03 & 0.551 & 0.415 & 0.290 \\
          & All*-ft. & \textbf{26.89} & \textbf{0.792} & \textbf{0.215} & 0.081 \\
          & Cat.-ft. & 25.67 & 0.760 & 0.238 & 0.099 \\
    \midrule
    \multirow{2}[2]{*}{pixelNeRF~\cite{AlexYu2021pixelNeRFNR}} & All* & 22.16 & 0.692 & 0.342 & 0.109 \\
          & Cat. & 20.65 & 0.676 & 0.348 & 0.195 \\
    \bottomrule
    \end{tabular}%
    }
  \label{tab:sparse_nerf_nvs}%
  \vspace{-15pt}
\end{table}%

\begin{figure*}[t]
    \centering
    \includegraphics[width=0.95\linewidth]{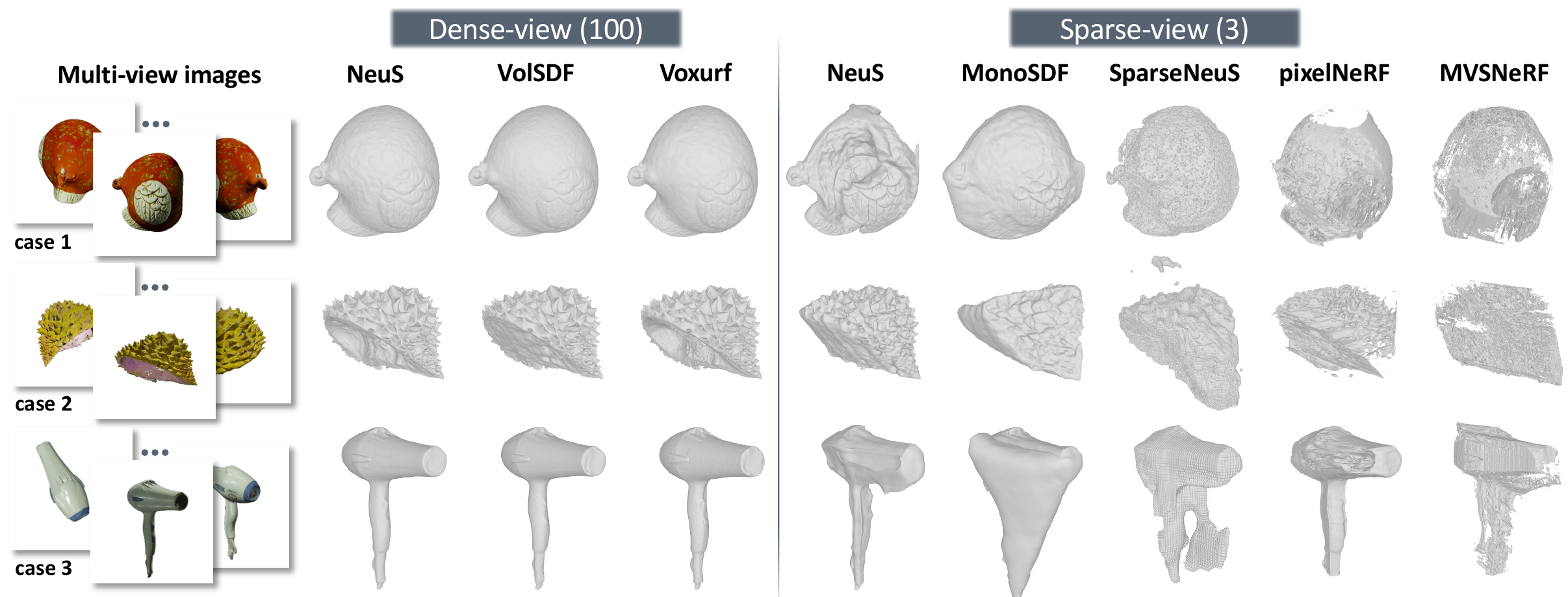}
    \caption{\small
    \textbf{Neural surface reconstruction results for both dense-view and sparse-view settings.}
    }
    \label{fig:sparse_surface}
    \vspace{-20pt}
\end{figure*} 

\noindent{\textbf{Single-Scene NVS.}}
We select three objects in each category for the experiments, randomly sampling 1/8 images as the hold-out test set. We involve NeRF~\cite{mildenhall2020nerf}, mip-NeRF~\cite{barron2021mip}, and a voxel-based system named Plenoxels~\cite{yu2021plenoxels} for comparison. As in Table~\ref{tab:single_nerf}, we find that Plenoxels achieve the best performance on average for PSNR, SSIM~\cite{wang2004image}, and LPIPS~\cite{zhang2018unreasonable}. There exists a clear margin for LPIPS between Plenoxels and the other two methods since voxel-based methods are especially good at modeling high-frequency appearance. We also present the standard deviation (SD) of results across all the training samples, where Plenoxels are relatively unstable compared to NeRF and mip-NeRF. We observe that Plenoxels introduce artifacts when encountering concave geometry (\eg, bowls, chairs) and suffer from an inaccurate density field modeling when the foreground object is dark. MLP-based methods are relatively robust against these difficult cases.
In a nutshell, our dataset provides a library with a variety of shapes and appearances, allowing a comprehensive evaluation of different NVS methods. See the supplementary for more results with the iPhone videos, detailed analysis, and visualizations. 

\noindent{\textbf{Cross-Scene NVS.}}
We conduct extensive experiments on novel view synthesis from sparse inputs by pixelNeRF~\cite{AlexYu2021pixelNeRFNR}, IBRNet~\cite{QianqianWang2021IBRNetLM} and MVSNeRF~\cite{AnpeiChen2021MVSNeRFFG} in Table~\ref{tab:sparse_nerf_nvs}. We select 10 categories with the most various scenes as the test set. All metrics are averaged over 300 images. In the generalization setting, although not trained on the test category, $\text{MVSNeRF}_{\text{All*}}$ is comparable to $\text{MVSNeRF}_{\text{Cat.}}$, $\text{IBRNet}_{\text{All*}}$ and $\text{pixelNeRF}_{\text{All*}}$ even outperforms the corresponding ``Cat''s in all terms of visual metrics, especially on regular-shaped objects such as squash and apple. It confirms that OmniObject3D serves as an information-rich dataset that is beneficial for obtaining a strong generalizable prior on unseen scenes. Moreover, it is noteworthy that MVSNeRF and pixelNeRF with `All*' generate better underlying depth than those with `Cat.', inferring generalizable methods can implicitly learn geometric cues though only trained from appearance in our dataset. It is reasonable that (1) IBRNet suffers more severely than the others in geometry under the scarcity of source context (only 3 views) as it is more suitable for dense-view generalization that complies with its view interpolation module. (2) MVSNeRF lags behind pixelNeRF on visual performance as we take 10 test frames widely distributed around the object in 360$^{\circ}$ by FPS sampling algorithm, where cost volume will be inaccurate on large-range viewpoint change. After further finetuning IBRNet for only around 10 minutes on each test scene, $\text{IBRNet}_{\text{All*-ft}}$ achieves the best view synthesis results, comparable to test-time optimized NVS methods on nearby views. It is promising to utilize the large-scale and category-prosperous OmniObject3D, to build a benchmark suite for evaluating diverse cross-scene NVS methods.


\begin{table}[t]
  \centering
  \small
  \setlength{\abovecaptionskip}{0mm}
  \caption{\textbf{Dense-view surface reconstruction results.}}
    \begin{tabular}{l||cccc}
    \toprule
    \multirow{2}[3]{*}{Method} & \multicolumn{4}{c}{Chamfer Distance $\times 10^3$ ($\downarrow$)} \\
\cmidrule{2-5}          & Hard  & Medium & Easy  & Avg \\
    \midrule
    NeuS~\cite{wang2021neus}  & 9.26  & 5.63  & 3.46  & 6.09 \\
    VolSDF~\cite{yariv2021volume} & 10.06 & \textbf{4.94}  & 2.86  & 5.92 \\
    Voxurf~\cite{wu2022voxurf} & \textbf{9.01}  & 4.98  & \textbf{2.58}  & \textbf{5.49} \\
    \midrule
    Avg   & 9.44  & 5.19  & 2.97  & 5.83 \\
    \bottomrule
    \end{tabular}%
  \label{tab:hardness_levels}%
  \vspace{-15pt}
\end{table}%

\begin{figure}[t]
    \centering
    \includegraphics[width=0.9\linewidth]{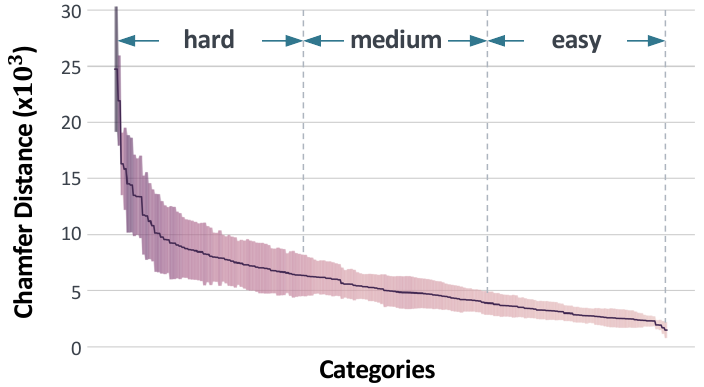}
    \setlength{\abovecaptionskip}{0mm}
    \caption{\small
    \textbf{Performance distribution of dense-view surface reconstruction.} The averaged results of the three methods is imbalanced. The colored area denotes a smoothed range of results. 
    }
    \label{fig:single_surface_levels}
    \vspace{-20pt}
\end{figure}

\subsection{Neural Surface Reconstruction}
\label{sec:surface}

Precise surface reconstruction from multi-view images enables a broad range of applications. 
For a single scene with dense-view images, algorithms are expected to conduct accurate, robust, and efficient surface reconstruction. 
When only sparse-view images are available, it is crucial to learn generalizable priors from a set of scenes or use other geometric cues to assist reconstruction.
Accordingly, we study the two settings for surface reconstruction methods.

\noindent{\textbf{Dense-View Surface Reconstruction.}}
Under this setting, we include three representative methods. NeuS~\cite{wang2021neus} and VolSDF~\cite{yariv2021volume} are two well-known approaches that bridge neural volume rendering with implicit surface representation. 
We also involve a voxel-based method called Voxurf~\cite{wu2022voxurf}, which leverages a hybrid representation to achieve acceleration and fine geometry reconstruction.  

Previous approaches in this task mainly perform evaluations and comparisons on 15 scenes from the DTU~\cite{aanaes2016large} dataset, which is not comprehensive and robust enough to demonstrate the ability of the methods in different scenarios. In comparison, we select three objects per category to run each of the three methods above, leading to over 1,500 reconstructions in total. We calculate the Chamfer Distance (CD) between the reconstructed surface and the ground truth. The distribution of the results is shown in Figure~\ref{fig:single_surface_levels}. 
The average curve is imbalanced: hard categories usually include low-textured,  concave, or complex shapes (\eg, bowls, vases, kennels, cabinets, and durians).
We thus split the categories into three levels of ``difficulty" based on the average curve, and the level-wise results are presented in Table~ \ref{tab:hardness_levels}. We can observe a clear margin among results in different levels for each method, indicating the split subsets to be generic and faithful.


\begin{table}[t]
  \centering
  \small
  \setlength{\abovecaptionskip}{0mm}
  \caption{\textbf{Sparse-view (3-view) surface reconstruction results.}}
  \setlength\tabcolsep{3.2pt}
    \begin{tabular}{l|c||cccc}
    \toprule
    \multirow{2}[3]{*}{Method} & \multirow{2}[3]{*}{Train} & \multicolumn{4}{c}{Chamfer Distance $\times 10^3$ ($\downarrow$)} \\
\cmidrule{3-6}          &       & Hard  & Medium & Easy  & Avg \\
    \midrule
    NeuS~\cite{wang2021neus} & Single & 29.35 & 27.62 & 24.79 & 27.3 \\
    MonoSDF~\cite{ZehaoYu2022MonoSDFEM} & Single & 35.14 & 35.35 & 32.76 & 34.68 \\
    \midrule
    \multirow{6}[2]{*}{SparseNeuS~\cite{XiaoxiaoLong2022SparseNeuSFG}} & 1 cat. & 34.05 & 31.32 & 31.14 & 32.36 \\
          & 10 cats. & 30.75 & 30.11 & 28.37 & 29.87 \\
          & All cats. & \textbf{26.13} & \textbf{26.08} & \textbf{22.13} & \textbf{25.00} \\
          & Easy  & 28.39 & 26.65 & 23.76 & 26.48 \\
          & Medium & 27.38 & 26.66 & 23.08 & 25.87 \\
          & Hard  & 27.42 & 26.95 & 24.63 & 26.47 \\
    \midrule
    MVSNeRF~\cite{AnpeiChen2021MVSNeRFFG} & All cats. & 56.68 & 48.09 & 48.70 & 51.16 \\
    pixelNeRF~\cite{AlexYu2021pixelNeRFNR} & All cats. & 63.31 & 59.91 & 61.47 & 61.56 \\
    \bottomrule
    \end{tabular}%
  \label{tab:sparse_surface}%
  \vspace{-20pt}
\end{table}%

\noindent{\textbf{Sparse-View Surface Reconstruction.}}
Dense-captured images of a scene are sometimes not available, so we also study the sparse-view scenario here. The following methods are included: 
NeuS~\cite{wang2021neus} with sparse-view input;
MonoSDF~\cite{ZehaoYu2022MonoSDFEM}, which takes in geometry cues estimated by pre-trained models;
SparseNeuS~\cite{XiaoxiaoLong2022SparseNeuSFG}, a generic surface prediction pipeline that learns generalizable priors; 
pixelNeRF~\cite{AlexYu2021pixelNeRFNR} and MVSNeRF~\cite{AnpeiChen2021MVSNeRFFG} from Sec.~\ref{sec:nerf}, whose geometries are extracted from the density field.
For NeuS, MonoSDF, and pixelNeRF, we use Farthest Point Sampling (FPS) to sample views that are most widely distributed; for SparseNeuS and MVSNeRF, we conduct FPS among the nearest 30 camera poses from a random reference view. We sample 3 views in all the experiments.

The quantitative and qualitative comparisons are shown in Table~\ref{tab:sparse_surface} and Figure~\ref{fig:sparse_surface}, respectively. We observe apparent artifacts in all the sparse-view reconstructed results. Among them, SparseNeuS trained on enough data demonstrates the best quantitative performance on average, and the pre-division on the train set does not result in a noticeable difference across difficulty levels. NeuS with sparse-view input achieves a surprisingly good performance. As shown in Figure~ \ref{fig:sparse_surface}, the FPS sampling equips NeuS with a coherent global shape for thin structures like case 3, while it encounters severe local geometry ambiguity like case 1. MonoSDF, on the contrary, partially overcomes the issue of ambiguity via the assistance of predicted geometry cues in case 1. However, it relies heavily on the accuracy of the estimated depth and normal and thus easily fails when the estimation is inaccurate (\eg, case 2 and 3). 
The surfaces extracted from generalized NeRF models, \ie, pixelNeRF and MVSNeRF, are of relatively low quality.

In brief, the challenging problem of sparse-view surface reconstruction has not been solved well currently. OmniObject3D is a promising database to study generalizable surface reconstruction pipelines as well as strategies for a robust usage of estimated geometry cues for this track.

\begin{figure}[t]
    \centering
    \includegraphics[width=1.0\linewidth]{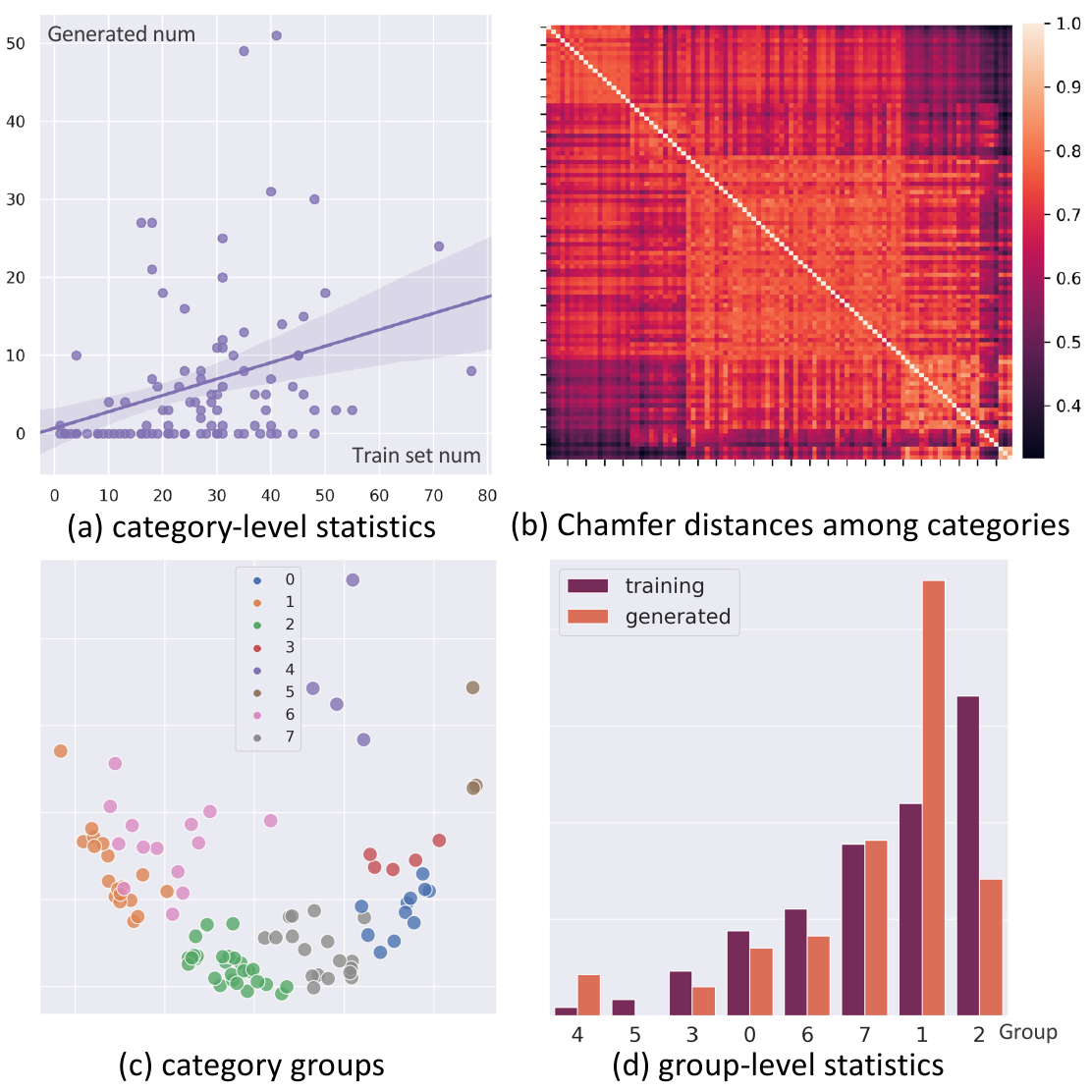}
        \setlength{\abovecaptionskip}{-3mm}
    \caption{\small
    \textbf{The category distribution of the generated shapes.} 
    \textbf{(a)} shows a weak positive correlation between the number of generated shapes and training shapes per category. 
    \textbf{(b)} visualizes the correlation matrix among different categories by Chamfer Distance between their mean shapes.  
    \textbf{(c)} visualizes categories being clustered into eight groups by KMeans. 
    \textbf{(d)} presents a clear training-generation relation in the group-level statistics.
    }
    \label{fig:gen_distribution}
    \vspace{-15pt}
\end{figure}

\begin{figure*}[t]
	\centering
	\includegraphics[width=1.0\linewidth]{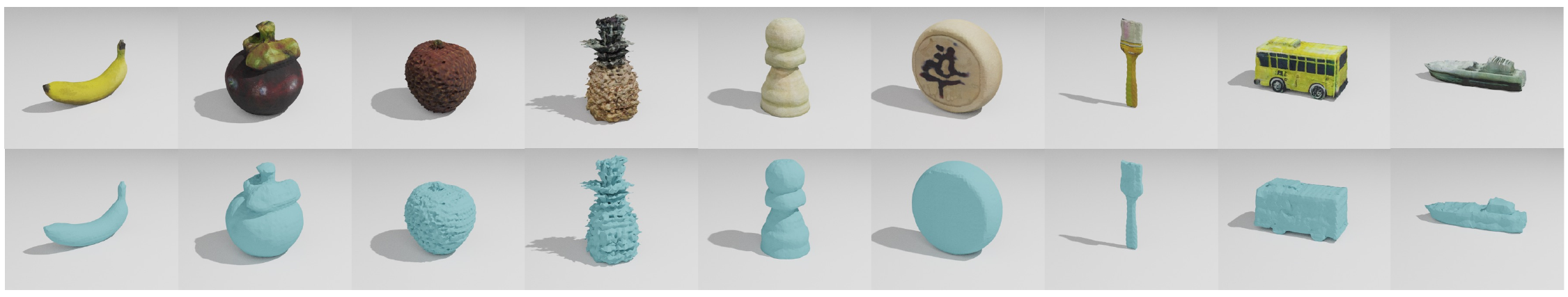}
        \setlength{\abovecaptionskip}{-3mm}
	\caption{\small
	\textbf{Examples of the generated textured shapes rendered in Blender.} OmniObject3D enables GET3D with realistic generations across a wide range of categories. 
	}
	\label{fig:get3d_examples}
    \vspace{-20pt}
\end{figure*}

\begin{figure}[t]
	\centering
	\includegraphics[width=1.0\linewidth]{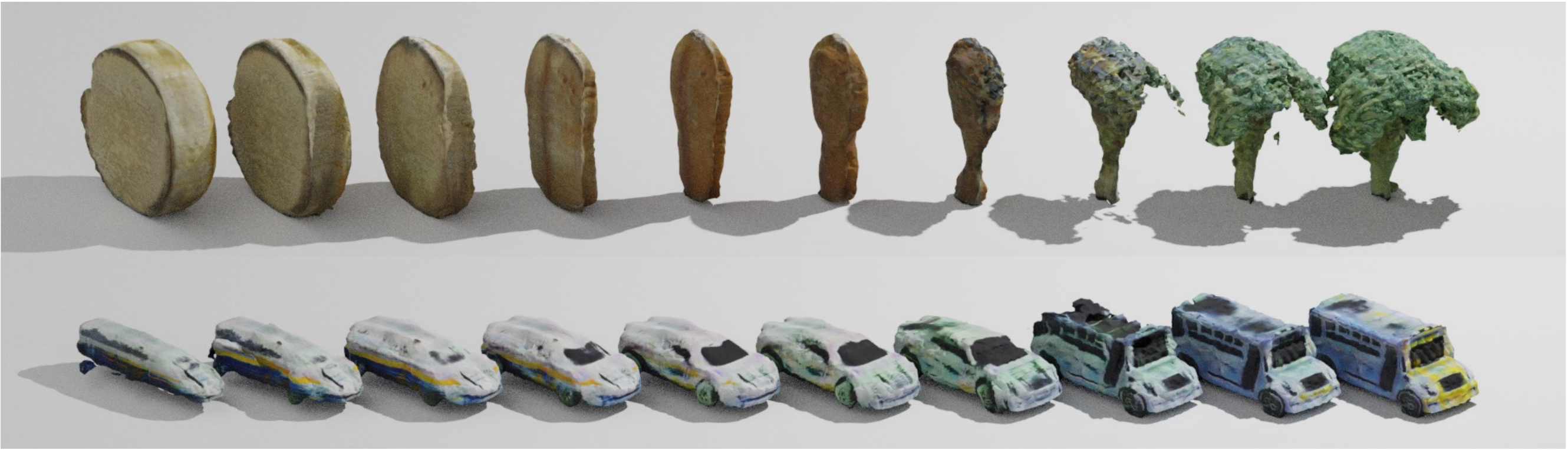}
        \setlength{\abovecaptionskip}{-3mm}
	\caption{\small
	\textbf{Shape interpolation.} We interpolate both geometry and texture latent codes from left to right.
	}
	\label{fig:interpolation}
	\vspace{-15pt}
\end{figure}

\subsection{3D Object Generation}
\label{sec:generation}
In this section, we adopt a state-of-the-art generative model that directly generates explicit textured 3D meshes, namely GET3D~\cite{gao2022get3d}. GET3D is originally evaluated on six categories (\textit{Car, Chair, Motorbike, Animal, House, and Human Body}) with independent models trained on each category. The number of shapes per category ranges from 337 to 7,497. In comparison, OmniObject3D contains many more categories with fewer objects in each. As a result, it is natural to train each model with multiple categories. 

We first provide some qualitative results in Figure~\ref{fig:get3d_examples}, where we show various generation results: 
The textures are rather realistic, and the shapes are coherent, enhanced by fine geometry details (\eg, the lychee and pineapple). 
We explore the latent space of the model and show interpolation results in Figure~\ref{fig:interpolation}. We can observe a smooth transition across instances that are semantically different.
We would like to further analyze the performance of the generative model trained on OmniObject3D from three aspects, \ie , \textit{semantic distribution}, \textit{diversity}, and \textit{quality}. 


\noindent{\textbf{Semantic Distribution.}} We randomly select 100 categories to train an unconditional model jointly. We randomly generate 1,000 textured meshes at inference time and ask human experts to label them. Shapes with ambiguity are not counted.  
Figure~\ref{fig:gen_distribution} (a) shows that the generated shapes per category are highly imbalanced, exhibiting a weak positive correlation with the training shape numbers. 
Actually, the categories are not independent but rather highly correlated. We calculate the ``mean shape'' for each category and visualize the Chamfer Distance among them in Figure~\ref{fig:gen_distribution} (b), which indicates that they can be further grouped.
Regarding each matrix row as a feature vector, we use KMeans to cluster them into eight groups (Figure~\ref{fig:gen_distribution} (c)) and carry out \textit{group-level statistics} in Figure~\ref{fig:gen_distribution} (d). It demonstrates a clear trend that the number of generated shapes increases along with or even faster than the number of training shapes in the group, revealing an enlarged semantic bias during generation. However, it also depends on the inner-group divergence. For example, Group 2 (883 shapes in 27 categories) has the largest number of training samples, while the high variation among its categories prevents it from dominating the generated shapes; Group 1 (587 shapes in 18 categories) has a relatively small divergence, which becomes the most popular in the generated shapes.
We present more details in the supplementary material.

\noindent{\textbf{Diversity and Quality.}} 
We select four representative data subsets for training and evaluation, namely \textit{fruits}, \textit{furniture}, \textit{toys}, and \textit{Rand-100}. We randomly split each subset into training (80\%) and testing (20\%).
We leverage three evaluation metrics: for geometry, we use Chamfer Distance (CD) to compute the Coverage score (Cov) and Minimum Matching Distance (MMD), which focus on the diversity and quality of the shapes, respectively; for texture, we adopt the widely-used FID~\cite{heusel2017gans}. But the FID metrics with different test splits are not directly comparable, suffering from a large variance when the test set is small. We thus introduce FID$^{\text{ref}}$ for reference, which is the FID between the train and test set. The results are shown in Table~\ref{tab:generation}. 
\textit{Furniture} suffers from the lowest quality (MMD) since the small train set with 17 categories is a difficult training source.
\textit{Fruits} has the same number of categories while being 2.3 times larger in scale, and some fruits share a very similar structure, leading to relatively higher quality and lower diversity (Cov).
\textit{Toys} achieve the best quality by training on only 7 categories.
\textit{Rand-100} is the most difficult case, and we can observe a trade-off between quality and diversity. Both the FID and FID$^{\text{ref}}$ are high for the first three subsets due of the small testing sets, while only `Rand-100' is relatively low.


\begin{table}[t]
  \centering
  \small
  \setlength{\abovecaptionskip}{0mm}
  \caption{\textbf{Quantitative evaluations on different data splits.}}
  \setlength\tabcolsep{3.2pt}
  \resizebox{.48\textwidth}{!} {
    \begin{tabular}{l|cc||cccc}
    \toprule
    Split & \#Objs & \#Cats & Cov (\% $ \uparrow$) & MMD ($\downarrow$) & FID ($\downarrow$) & FID$^{\text{ref}}$ \\
    \midrule
    Furniture & 265   & 17    & \textbf{67.92} & 4.27  & 87.39 & 58.40 \\
    Fruits & 610   & 17    & 46.72 & 3.32  & 105.31 & 87.15 \\
    Toys  & 339   & 7     & 55.22 & \textbf{2.78}  & 122.77 & 41.40 \\
    Rand-100 & 2951  & 100   & 61.70 & 3.89  & \textbf{46.57} & \textbf{8.65} \\
    \bottomrule
    \end{tabular}%
    }
  \label{tab:generation}%
  \vspace{-15pt}
\end{table}%

In brief, training and evaluating generative models on a large-vocabulary and realistic dataset is a promising but challenging task. We reveal crucial problems like the semantic distribution bias and varying exploration difficulties in different groups. OmniObject3D serves as a great database for further examination in this area.
\section{Conclusion and Outlook}
We introduce OmniObject3D, a large vocabulary 3D object dataset with massive high-quality real-scanned 3D objects, including 6,000 objects from 190 categories. It provides rich annotations, including textured 3D meshes, sampled point clouds, posed multi-view images rendered by Blender, and real-captured video frames with foreground masks and COLMAP camera poses. We set up four evaluation tracks, revealing new observations, challenges, and opportunities for future research in realistic 3D vision.


We will regulate the usage of our data to avoid potential negative social impacts.

\noindent\textbf{Acknowledgement.}
This project is funded by Shanghai AI Laboratory, CUHK Interdisciplinary AI Research Institute, the Centre for Perceptual and Interactive Intelligence (CPIl) Ltd under the Innovation and Technology Commission (ITC)'s InnoHK, Hong Kong RGC Theme-based Research Scheme 2020/21 (No. T41-603/20- R), OpenXDLab, the Ministry of Education, Singapore, under its MOE AcRF Tier 2 (MOE-T2EP20221-0012), NTU NAP, and under the RIE2020 Industry Alignment Fund – Industry Collaboration Projects (IAF-ICP) Funding Initiative.

\appendix

\setcounter{table}{0}
\setcounter{figure}{0}
\renewcommand{\thetable}{R\arabic{table}}
\renewcommand\thefigure{S\arabic{figure}}
\section{Additional Information of OmniObject3D}
\label{supp:dataset}
\begin{figure}[t]
    \centering
    \includegraphics[width=0.8\linewidth]{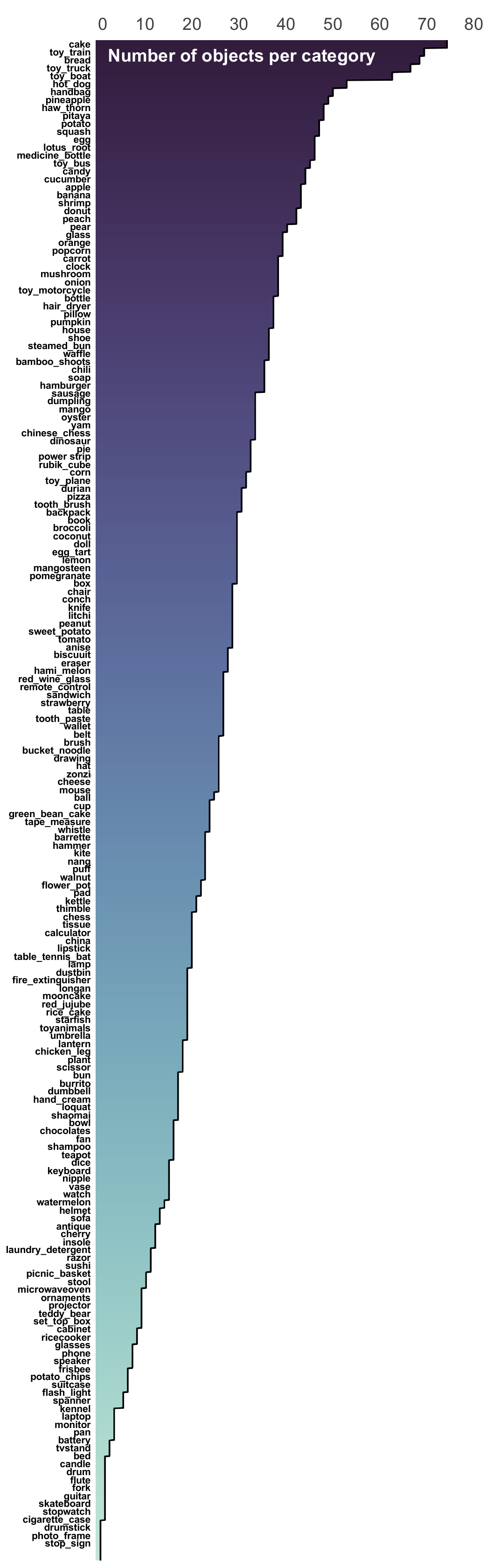}
    \setlength{\abovecaptionskip}{0mm}
    \caption{\small
    \textbf{A full class list with number of objects per category.}
    }
    \label{fig:full_list}
    \vspace{-10pt}
\end{figure}

\begin{figure}[t]
    \centering
    \includegraphics[width=0.9\linewidth]{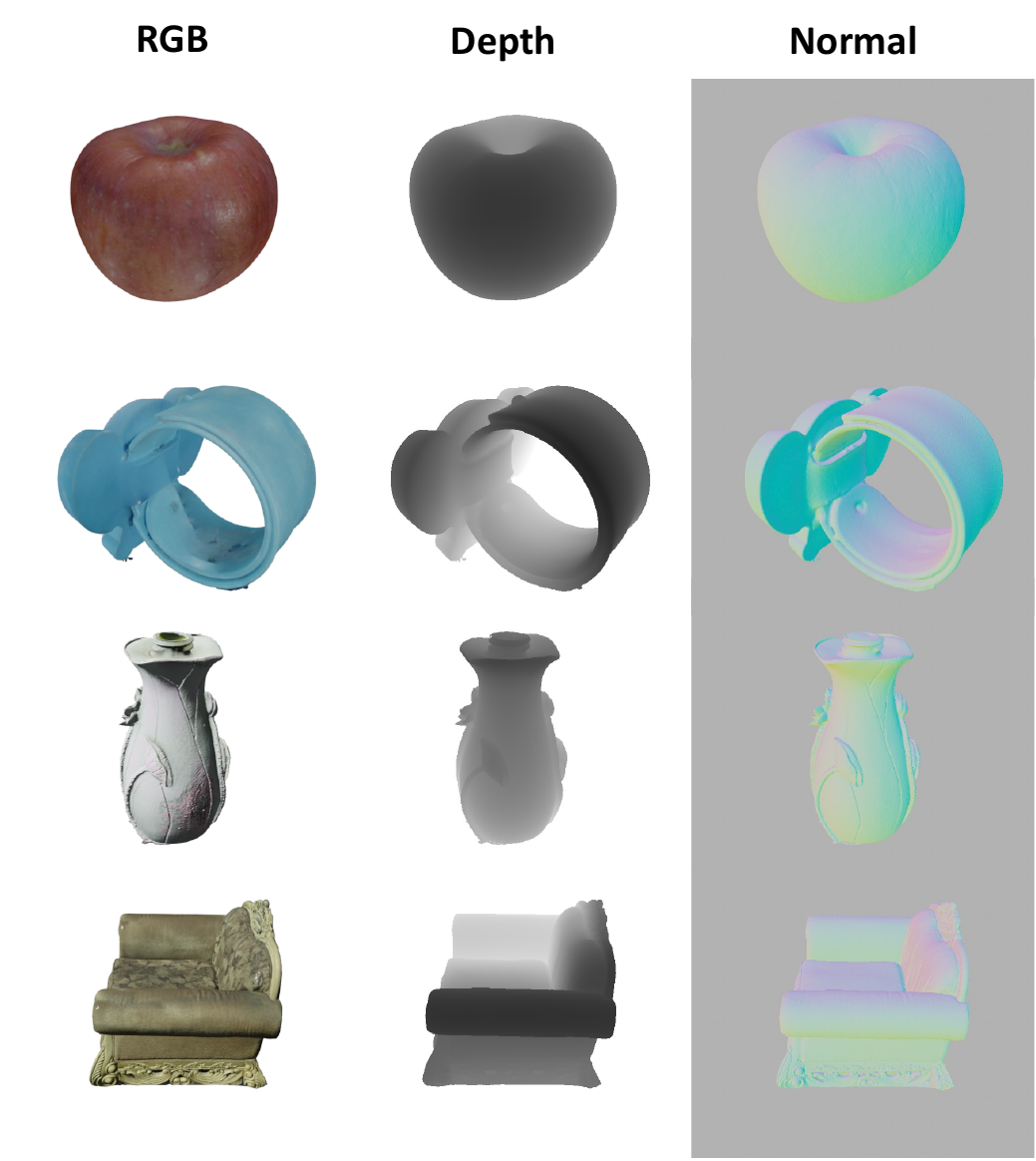}
    \setlength{\abovecaptionskip}{0mm}
    \caption{\small
    \textbf{Examples of the Blender~\cite{blender} rendered results.}
    }
    \label{fig:rgb_depth_normal}
    \vspace{-10pt}
\end{figure}

\begin{figure}[t]
    \centering
    \includegraphics[width=1.\linewidth]{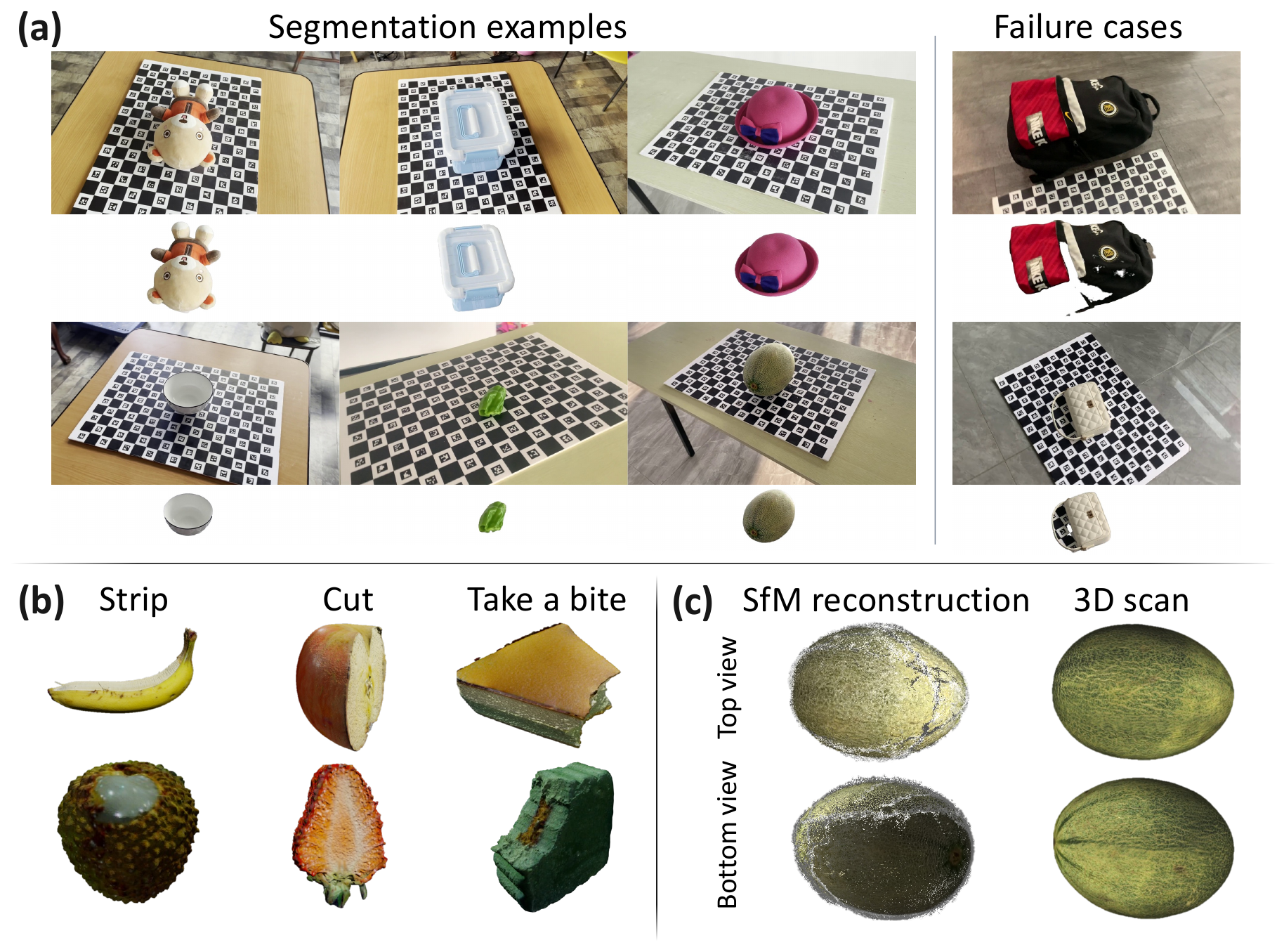}
    \setlength{\abovecaptionskip}{0mm}
    \caption{\small
    \textbf{Examples of the segmentation (a), manipulation (b), and reconstruction (c).} In (c), the missing bottom of the SfM reconstruction from video frames is due to its touch with the table. 
    }
    \label{fig:more_visualization}
    \vspace{-15pt}
\end{figure}

We first provide a full category list with the number of objects for each class in Figure~\ref{fig:full_list}. Most of the categories have [10, 40] objects. The dataset includes objects that have undergone common manipulations, as shown in Figure~\ref{fig:more_visualization} (b).
For each object, the raw data includes a textured 3D mesh and several surrounding videos. To demonstrate the completeness and high quality of our scanned objects, we compare the quality between the COLMAP sparse reconstruction and the textured mesh from the scanner in Figure~\ref{fig:more_visualization} (c).
Given a high-fidelity 3D scan, we can render realistic and high-resolution multi-view images with modern graphics engines like the Blender~\cite{blender}, where we also save the corresponding depth and normal maps (Figure~\ref{fig:rgb_depth_normal}) for different research usage.
We also provide the users with posed frames from the real-captured videos following \cite{reizenstein2021co3d}. We leverage the calibration board and COLMAP~\cite{schonberger2016sfm} to recover the poses of selected frames with a real-world scale, as described in the main text, and then we develop a matting pipeline based on a two-stage U$^2$-Net~\cite{Qin_2020_PR} model together with a post-processing FBA~\cite{forte2020fbamatting} model. In detail, we first utilize the Rembg~\footnote{https://github.com/danielgatis/rembg} tool on image frames to remove backgrounds from different categories and choose 3,000 good results as the pseudo segmentation labels. We then refine our pipeline by fine-tuning with the pseudo labels to boost its segmentation ability on objects. We show some examples and failure cases of our segmentation pipeline in Figure~\ref{fig:more_visualization} (a).

\section{Related Works}
\label{supp:related_works}
We have briefly discussed the related works for the four benchmarks in the main text, and we conduct a more comprehensive discussion here.

\noindent{\textbf{Robust Point Cloud Perception.}}
Robustness to out-of-distribution (OOD) data has been an important topic in point cloud perception since point clouds are widely employed in safety-critical applications, \eg, autonomous driving. In particular,  OOD styles (\eg, different styles in CAD models and real-world objects) and OOD corruptions (\eg, missing points) are two main challenges to point cloud perception robustness. A line of work~\cite{qi2016pointnet, wang2019dgcnn, chen2020pointmixup, kim2021pointwolf} evaluates the robustness to OOD corruptions by adding corruptions like random jittering and rotation to clean test sets. Recent work works~ \cite{taghanaki2020robustpointset, ren2022modelnet-c} further systematically anatomize the corruptions and propose a standard corruption test suite. However, they fail to take account of OOD styles. Another line of work~\cite{reizenstein2021co3d,ahmadyan2021objectron} evaluates the sim-to-real domain gap by testing models trained on clean synthetic datasets (\eg, ModelNet-40~\cite{wu20153d}) on noisy real-world test sets (\eg, ScanObjectNN~\cite{uy2019revisiting}). However, the sim-to-real gap couples OOD styles and OOD corruptions at the same time, which makes the results hard to analyze. In this work, we use OmniObject3D dataset to provide high-quality real-world point cloud to measure the OOD style robustness, and apply systematic corruptions on top of it to measure the OOD corruptions robustness. We hence provide the first point cloud perception benchmark that allows fine-grained evaluation of the robustness on both OOD styles and corruptions.

\noindent{\textbf{Neural Radiance Field.}}
Neural radiance field (NeRF)~\cite{mildenhall2020nerf} represents a scene with a fully-connected deep network (MLPs), which takes in hundreds of sampled points along each camera ray and outputs the predicted color and density. Novel views of the scene are synthesized by projecting the colors and densities into an image via volume rendering.
Inspired by the success of NeRF, a massive follow-up effort has been made to improve its quality~\cite{barron2021mip, DorVerbin2022RefNeRFSV, BenMildenhall2021NeRFIT, JonathanTBarron2021MipNeRF3U}, and efficiency~\cite{yu2021plenoxels, sun2021direct, chen2022tensorf, mueller2022instant}.
A branch of works~\cite{QianqianWang2021IBRNetLM, AlexYu2021pixelNeRFNR, AnpeiChen2021MVSNeRFFG, reizenstein2021co3d, YuanLiu2021NeuralRF} has also explored the generalization ability of NeRF-based frameworks. 
PixelNeRF~\cite{AlexYu2021pixelNeRFNR}, MVSNeRF~\cite{AnpeiChen2021MVSNeRFFG}, IBRNet~\cite{QianqianWang2021IBRNetLM}, and NeuRay~\cite{YuanLiu2021NeuralRF} reconstruct the radiance field with a mere forward pass during inference via training on cross-scenes.
NeRFormer~\cite{reizenstein2021co3d}, IBRNet~\cite{QianqianWang2021IBRNetLM}, and GNT~\cite{MukundVarma2022IsAA} leverage Transformers for generalizable NeRF.

\begin{table*}
\centering\small
\caption{\textbf{Point cloud perception robustness analysis on OmniObject3D with different architecture designs.} Models are trained on ModelNet-40 dataset. OA on OmniObject3D evaluates the robustness to OOD styles. mean Corruption Error (mCE) on the corrupted OmniObject3D-C evaluates the robustness to OOD corruptions. The \textcolor{blue}{blue} cells denote best in each row, and the \textcolor{red}{red} cells denote the worst.}
\label{tab:pcd_robustness_full}
\begin{tabular}{lc|c||ccccccc|c}
\toprule
{} &          $\text{OA}_\text{Clean}\uparrow$ &                               $\text{OA}_\text{Style}\uparrow$ &                                       Scale &                                      Jitter &                                      Drop-G &                                      Drop-L &                                       Add-G &                                       Add-L &                                      Rotate &                                     $\text{mCE}\downarrow$ \\
\midrule
DGCNN~\cite{wang2019dgcnn}      & 
\cellcolor[HTML]{FFFFFF}0.926 &
\cellcolor[HTML]{FFFFFF}0.448 &
\cellcolor[HTML]{FFFFFF}1.000 &               \cellcolor[HTML]{FFFFFF}1.000 &               \cellcolor[HTML]{FFFFFF}1.000 &               \cellcolor[HTML]{FFFFFF}1.000 &               \cellcolor[HTML]{FFFFFF}1.000 &               \cellcolor[HTML]{FFFFFF}1.000 &               \cellcolor[HTML]{FFFFFF}1.000 &               \cellcolor[HTML]{FFFFFF}1.000 \\
PointNet~\cite{qi2016pointnet}   &
\cellcolor[HTML]{FFFFFF}0.907 &
\cellcolor[HTML]{FFFFFF}0.466 &  \cellcolor[HTML]{FFFFFF} \underline{0.925} &     \cellcolor[HTML]{FFFFFF} \textbf{0.858} &               \cellcolor[HTML]{FFFFFF}0.976 &     \cellcolor[HTML]{DFE7FD} \textbf{0.816} &               \cellcolor[HTML]{FDE2E4}1.318 &               \cellcolor[HTML]{FFFFFF}0.921 &               \cellcolor[HTML]{FFFFFF}0.935 &               \cellcolor[HTML]{FFFFFF}0.969 \\
PointNet++~\cite{qi2017pointnetplusplus}  & 
\cellcolor[HTML]{FFFFFF}0.930 &
\cellcolor[HTML]{FFFFFF}0.407 &               \cellcolor[HTML]{FFFFFF}1.104 &               \cellcolor[HTML]{FFFFFF}1.071 &               \cellcolor[HTML]{FFFFFF}1.108 &               \cellcolor[HTML]{DFE7FD}0.886 &               \cellcolor[HTML]{FFFFFF}1.101 &               \cellcolor[HTML]{FDE2E4}1.123 &               \cellcolor[HTML]{FFFFFF}1.031 &               \cellcolor[HTML]{FFFFFF}1.066 \\
RSCNN~\cite{liu2019rscnn}      &
\cellcolor[HTML]{FFFFFF}0.923 &
\cellcolor[HTML]{FFFFFF} 0.393 &               \cellcolor[HTML]{FFFFFF}1.115 &               \cellcolor[HTML]{FFFFFF}1.078 &               \cellcolor[HTML]{FDE2E4}1.144 &               \cellcolor[HTML]{DFE7FD}0.997 &               \cellcolor[HTML]{FFFFFF}1.042 &               \cellcolor[HTML]{FFFFFF}1.079 &               \cellcolor[HTML]{FFFFFF}1.025 &               \cellcolor[HTML]{FFFFFF}1.076 \\
SimpleView~\cite{goyal2021simpleview} &
\cellcolor[HTML]{FFFFFF}\textbf{0.939} &
\cellcolor[HTML]{FFFFFF}0.476 &               \cellcolor[HTML]{DFE7FD}0.940 &               \cellcolor[HTML]{FFFFFF}0.951 &               \cellcolor[HTML]{FFFFFF}0.959 &               \cellcolor[HTML]{FFFFFF}1.012 &               \cellcolor[HTML]{FDE2E4}1.043 &               \cellcolor[HTML]{FFFFFF}1.037 &               \cellcolor[HTML]{FFFFFF}0.949 &               \cellcolor[HTML]{FFFFFF}0.990 \\
GDANet~\cite{xu2021gdanet}     &
\cellcolor[HTML]{FFFFFF}0.934 &
\cellcolor[HTML]{FFFFFF}\underline{0.497} &     \cellcolor[HTML]{FFFFFF} \textbf{0.887} &               \cellcolor[HTML]{FFFFFF}0.933 &  \cellcolor[HTML]{FFFFFF} \underline{0.923} &               \cellcolor[HTML]{FDE2E4}0.975 &     \cellcolor[HTML]{FFFFFF} \textbf{0.884} &               \cellcolor[HTML]{FFFFFF}0.921 &     \cellcolor[HTML]{DFE7FD} \textbf{0.882} &     \cellcolor[HTML]{FFFFFF} \textbf{0.920} \\
PAConv~\cite{xu2021paconv}     &
\cellcolor[HTML]{FFFFFF}0.936 &
\cellcolor[HTML]{FFFFFF} 0.403 &               \cellcolor[HTML]{FFFFFF}1.034 &               \cellcolor[HTML]{FFFFFF}1.101 &               \cellcolor[HTML]{DFE7FD}1.032 &               \cellcolor[HTML]{FFFFFF}1.052 &               \cellcolor[HTML]{FDE2E4}1.159 &               \cellcolor[HTML]{FFFFFF}1.057 &               \cellcolor[HTML]{FFFFFF}1.082 &               \cellcolor[HTML]{FFFFFF}1.073 \\
CurveNet~\cite{xiang2021curvenet}   &
\cellcolor[HTML]{FFFFFF}\underline{0.938} &
\cellcolor[HTML]{FFFFFF}\textbf{0.500} &               \cellcolor[HTML]{FFFFFF}0.930 &  \cellcolor[HTML]{FFFFFF} \underline{0.930} &     \cellcolor[HTML]{FFFFFF} \textbf{0.920} &               \cellcolor[HTML]{DFE7FD}0.869 &               \cellcolor[HTML]{FFFFFF}0.929 &               \cellcolor[HTML]{FDE2E4}0.997 &  \cellcolor[HTML]{FFFFFF} \underline{0.907} &  \cellcolor[HTML]{FFFFFF} \underline{0.929} \\
PCT~\cite{guo2020pct}        &
\cellcolor[HTML]{FFFFFF}0.930 &
\cellcolor[HTML]{FFFFFF}0.459 &               \cellcolor[HTML]{FFFFFF}0.950 &               \cellcolor[HTML]{FFFFFF}0.986 &               \cellcolor[HTML]{FDE2E4}1.011 &               \cellcolor[HTML]{DFE7FD}0.862 &  \cellcolor[HTML]{FFFFFF} \underline{0.921} &  \cellcolor[HTML]{FFFFFF} \underline{0.912} &               \cellcolor[HTML]{FFFFFF}1.001 &               \cellcolor[HTML]{FFFFFF}0.940 \\
RPC~\cite{ren2022modelnet-c}        &
\cellcolor[HTML]{FFFFFF}0.930 &
\cellcolor[HTML]{FFFFFF}0.472 &               \cellcolor[HTML]{FFFFFF}0.947 &               \cellcolor[HTML]{FFFFFF}0.940 &               \cellcolor[HTML]{FFFFFF}0.967 &  \cellcolor[HTML]{DFE7FD} \underline{0.855} &               \cellcolor[HTML]{FDE2E4}0.999 &     \cellcolor[HTML]{FFFFFF} \textbf{0.909} &               \cellcolor[HTML]{FFFFFF}0.915 &               \cellcolor[HTML]{FFFFFF}0.936 \\
\bottomrule
\end{tabular}
\vspace{-10pt}
\end{table*}

\begin{table}[t]
\centering
\small
\caption{\textbf{Comparisons of 3 single-scene NVS methods on different data types.} 
For all the methods we involve, we can observe that the \emph{Blender} setting performs the best; the \emph{SfM-wo-bg} setting is a little bit worse due to the motion blur and potential inaccuracy in SfM pose estimation; the \emph{SfM-w-bg} setting always achieves the lowest PSNR, as the background in the unbounded scene introduces further challenges.}
\begin{tabular}{c|c|c}
\toprule
Method                     & Data-type                     &  PSNR ($\uparrow$)\\ 
\midrule 
\multirow{3}{*}{NeRF~\cite{mildenhall2020nerf}}  
    & SfM-w-bg       & 22.92 \\ 
    & SfM-wo-bg       & 24.70 \\ 
    & Blender      & \textbf{28.07}       \\ 
\midrule
\multirow{3}{*}{Mip-NeRF~\cite{barron2021mip}}  
    & SfM-w-bg      & 23.29 \\ 
    & SfM-wo-bg       & 25.62  \\ 
    & Blender     & \textbf{31.25}    \\ 
\midrule
\multirow{3}{*}{Plenoxel~\cite{yu2021plenoxels}}  
    & SfM-w-bg      & 14.06 \\ 
    & SfM-wo-bg       & 19.18 \\ 
    & Blender    & \textbf{28.07}   \\ 
\bottomrule
\end{tabular}
  \label{tab:single_nerf_scenario}
  \vspace{-10pt}
\end{table}

\noindent{\textbf{Neural Surface Reconstruction.}}
Implicit Neural Representations (INR)~\cite{park2019deepsdf,chen2019learning,lombardi2019neural,mescheder2019occupancy,sitzmann2019srns,saito2019pifu,atzmon2019controlling,jiang2020sdfdiff,zhang2021learning,toussaint2022hal} of 3D object geometry and appearance with neural networks have attracted increasing attention in recent years. 
Some approaches \cite{niemeyer2020differentiable,yariv2020multiview,liu2020dist,kellnhofer2021neural} regard the color of an intersection point between the ray and the surface as the rendered color, namely surface rendering, and they typically rely on accurate object masks.
Another trend of recent approaches~\cite{oechsle2021unisurf,yariv2021volume,wang2021neus,francois2021warping,wu2022voxurf} proposes to leverage neural radiance field with implicit surface representations like Signed Distance Function (SDF) for higher-quality and mask-free surface reconstruction from multi-view images.
NeuS~\cite{wang2021neus}, VolSDF~\cite{yariv2021volume} reconstruct implicit surfaces with an SDF-based volume rendering scheme, and Voxurf~\cite{wu2022voxurf} leverages an explicit volumetric representation for acceleration.
Since dense camera views of scenes are sometimes unavailable, SparseNeuS~\cite{XiaoxiaoLong2022SparseNeuSFG} and MonoSDF~\cite{ZehaoYu2022MonoSDFEM} explore surface reconstruction from sparse views. The former exploits generalizable priors cross scenes for a generic surface prediction, while the latter takes advantage of the estimated geometry cues predicted by pretrained networks.

OmniObject3D can serve as a large-scale benchmark with realistic photos and meshes for both training and evaluation. It bears a large vocabulary and high diversity in shape and appearance, offering an opportunity for pursuing more generalizable and robust novel view synthesis and surface reconstruction methods. 

\begin{figure*}[t]
    \centering
    \includegraphics[width=0.95\linewidth]{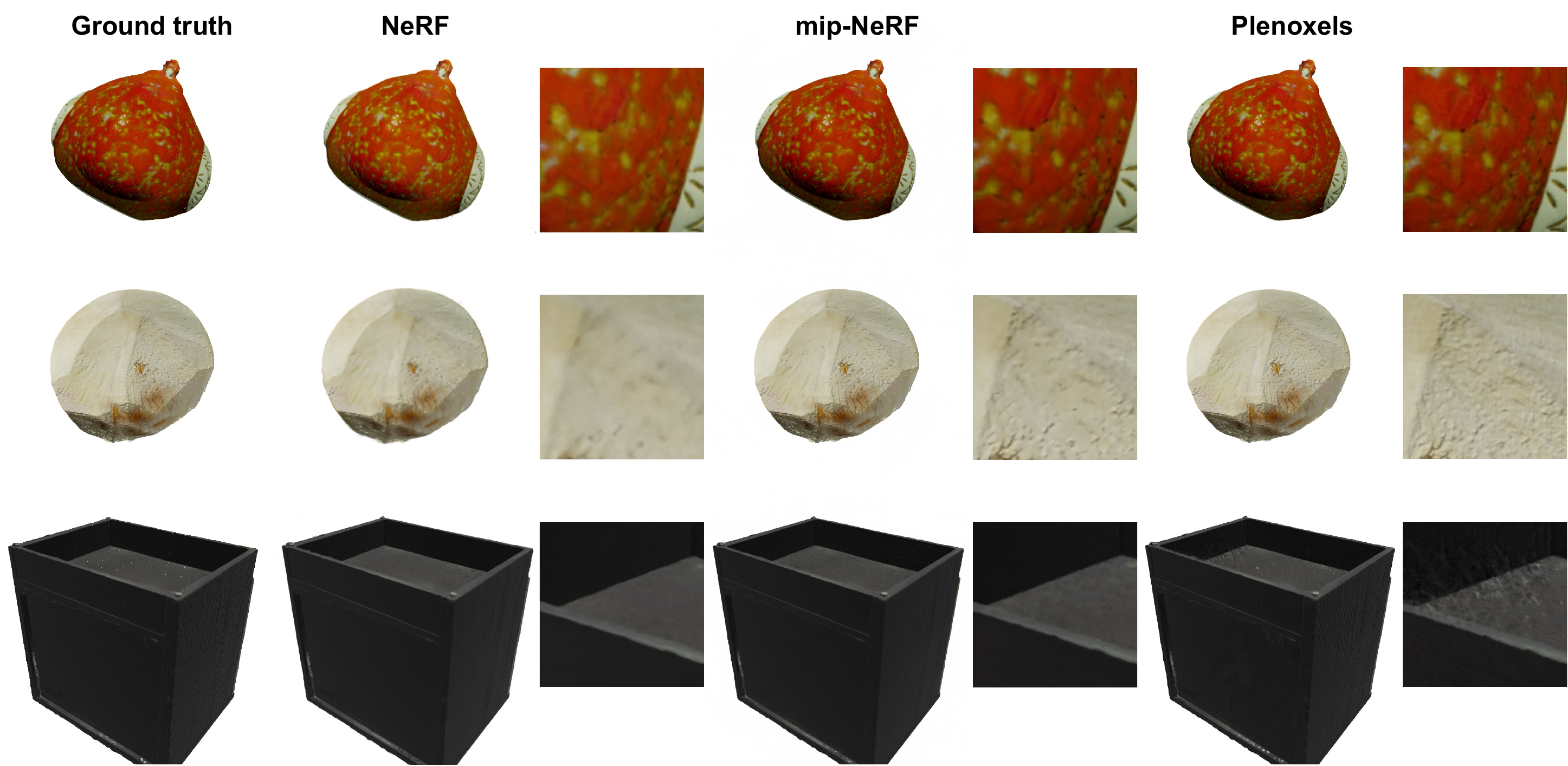}
    \caption{\small
    \textbf{Qualitative comparisons of single-scene NVS methods in different rendered scenes from our dataset.}
    }
    \label{fig:single_nvs}
\end{figure*}

\noindent{\textbf{3D Object Generation.}}
Recent advances in photorealistic 2D image generations ~\cite{TeroKarras2018ASG,LehtinenJaakko2019AnalyzingAI,TeroKarras2021AliasFreeGA,TaesungPark2019SemanticIS,XunHuang2022MultimodalCI,OmmerBjrn2020TamingTF,PrafullaDhariwal2021DiffusionMB} inspire the explorations of 3D content generation. Early approaches~\cite{JiajunWu2016LearningAP,MatheusGadelha20163DSI,PhilippHenzler2018EscapingPC,SebastianLunz2020InverseGG,EdwardJSmith2017ImprovedAS} extend 2D generation frameworks to 3D voxels with a high computational cost when generating high-resolution contents. Some other works adopt different 3D data formulations, \eg, point cloud~\cite{PanosAchlioptas2017LearningRA,GuandaoYang2019PointFlow3P,LinqiZhou20213DSG,KaichunMo2019StructureNetHG} and octree~\cite{MoritzIbing2022OctreeTA} to generate coarse geometry. 
OccNet~\cite{LarsMescheder2018OccupancyNL}, IM-NET~\cite{ZhiqinChen2018LearningIF} generates the 3D meshes with implicit representation while extracting high-quality surfaces is non-trivial. Encouraged by NeRF~\cite{mildenhall2020nerf}, extensive works~\cite{EricRChan2020piGANPI,KatjaSchwarz2020GRAFGR,MichaelNiemeyer2020GIRAFFERS,ZekunHao2021GANcraftU3,JiataoGu2021StyleNeRFAS,PengZhou2021CIPS3DA3,EricRChan2022EfficientG3,RoyOrEl2022StyleSDFH3,KatjaSchwarz2022VoxGRAFF3,YinghaoXu20223DawareIS} explore 3D-aware image synthesis rather than mesh generation. Aiming at generating textured 3D meshes, Textured3DGAN~\cite{DarioPavllo2021LearningGM} and DIBR~\cite{WenzhengChen2019LearningTP} deform template meshes, preventing them from complex shapes. PolyGen~\cite{CharlieNash2020PolyGenAA}, SurfGen~\cite{AndrewLuo2022SurfGenA3}, and GET3D~\cite{JunGao2022GET3DAG} generate meshes with arbitrary topology. Distinguishable from others, GET3D generates diverse meshes with rich geometry and textures. With the proposed OmniObject3D dataset, we extend the benchmarks of realistic 3D generation approaches to large vocabulary and massive objects, enabling the exploration of better generation quality and diversity.

\section{Additional Experimental Results}
\label{supp:experiments}
\subsection{Robust 3D Perception}
\label{supp:sec:pointcloud}

Following ModelNet-C~\cite{ren2022modelnet-c}, we perform seven kinds of out-of-distribution (OOD) corruptions for study, including “Scale”, “Jitter”, “Drop Global/Local”, “Add Global/Local”, and “Rotate”. Please refer to their paper for a detailed illustration of each corruption type.
We calculate the error under each corruption and the mean Corruption Error (mCE) is an average of the results. The full evaluation results corresponding to are shown in Table~\ref{tab:pcd_robustness_full}.

\subsection{Novel View Synthesis}
\subsubsection{Single-Scene NVS}

\begin{table*}[t]
\centering\small
\caption{\textbf{Cross-scene novel view synthesis results on 10 categories.} We evaluate our benchmarks on 3 unseen scenes per category with 3 source views. In each scene, we take 10 test frames widely distributed around the object by FPS sampling strategy.}
  \resizebox{.98\textwidth}{!} {
\begin{tabular}{c|c|c|cccccccccc}
\toprule
Method                     & Train                    & Metric   & toy train & bread & cake  & toy boat & hot dog & wallet & pitaya & squash & handbag & apple \\ \midrule
\multirow{16}{*}{MVSNeRF~\cite{AnpeiChen2021MVSNeRFFG}}  & \multirow{4}{*}{All*}    & PSNR     & 15.90     & 16.80 & 15.47 & 16.28    & 15.84   & 20.58  & 18.69  & 17.81  & 18.02   & 19.55 \\ 
                           &                          & SSIM     & 0.501     & 0.548 & 0.522 & 0.519    & 0.497   & 0.534  & 0.490  & 0.576  & 0.564   & 0.681 \\ 
                           &                          & LPIPS    & 0.480     & 0.456 & 0.480 & 0.408    & 0.429   & 0.449  & 0.456  & 0.417  & 0.444   & 0.403 \\ 
                           &                          &    $\mathcal{L}_{1}^{\text{depth}}$  & 0.182     & 0.155 & 0.249 & 0.253    & 0.127   & 0.261  & 0.178  & 0.187  & 0.229   & 0.113 \\ \cline{2-13} 
                           & \multirow{4}{*}{Cat.}    & PSNR     & 16.14     & 16.87 & 14.60 & 15.65    & 16.64   & 20.76  & 19.09  & 16.97  & 18.35   & 20.40 \\ 
                           &                          & SSIM     & 0.515     & 0.560 & 0.527 & 0.444    & 0.520   & 0.524  & 0.505  & 0.548  & 0.575   & 0.709 \\ 
                           &                          & LPIPS    & 0.475     & 0.463 & 0.488 & 0.433    & 0.431   & 0.464  & 0.449  & 0.435  & 0.444   & 0.399 \\ 
                           &                          &    $\mathcal{L}_{1}^{\text{depth}}$  & 0.175     & 0.127 & 0.339 & 0.477    & 0.134   & 0.382  & 0.237  & 0.101  & 0.219   & 0.112 \\ \cline{2-13} 
                           & \multirow{4}{*}{All*-ft} & PSNR     & 23.16     & 25.82 & 25.14 & 23.47    & 23.91   & 27.83  & 25.36  & 25.68  & 26.09   & 30.53 \\ 
                           &                          & SSIM     & 0.717     & 0.769 & 0.745 & 0.736    & 0.714   & 0.739  & 0.710  & 0.761  & 0.803   & 0.845 \\ 
                           &                          & LPIPS    & 0.281     & 0.224 & 0.263 & 0.228    & 0.248   & 0.293  & 0.227  & 0.255  & 0.280   & 0.215 \\ 
                           &                          &    $\mathcal{L}_{1}^{\text{depth}}$  & 0.091     & 0.062 & 0.081 & 0.141    & 0.053   & 0.078  & 0.069  & 0.061  & 0.130   & 0.053 \\ \cline{2-13} 
                           & \multirow{4}{*}{Cat.-ft} & PSNR     & 22.88     & 25.58 & 25.29 & 23.80    & 23.44   & 27.38  & 25.46  & 25.40  & 25.94   & 30.06 \\ 
                           &                          & SSIM     & 0.721     & 0.758 & 0.748 & 0.733    & 0.698   & 0.722  & 0.715  & 0.759  & 0.803   & 0.840 \\ 
                           &                          & LPIPS    & 0.283     & 0.243 & 0.262 & 0.226    & 0.280   & 0.318  & 0.229  & 0.277  & 0.283   & 0.244 \\ 
                           &                          &    $\mathcal{L}_{1}^{\text{depth}}$  & 0.122     & 0.053 & 0.064 & 0.096    & 0.060   & 0.084  & 0.071  & 0.048  & 0.120   & 0.046 \\ \midrule
\multirow{16}{*}{IBRNet~\cite{QianqianWang2021IBRNetLM}}   & \multirow{4}{*}{All*}    & PSNR     & 17.90     & 19.08 & 17.09 & 17.89    & 17.77   & 23.13  & 20.11  & 20.25  & 18.36   & 22.36 \\ 
                           &                          & SSIM     & 0.526     & 0.599 & 0.538 & 0.530    & 0.516   & 0.579  & 0.511  & 0.632  & 0.530   & 0.726 \\ 
                           &                          & LPIPS    & 0.430     & 0.383 & 0.422 & 0.368    & 0.394   & 0.426  & 0.405  & 0.356  & 0.451   & 0.352 \\ 
                           &                          &    $\mathcal{L}_{1}^{\text{depth}}$  & 0.379     & 0.327 & 0.610 & 0.357    & 0.338   & 0.419  & 0.388  & 0.392  & 0.847   & 0.175 \\ \cline{2-13} 
                           & \multirow{4}{*}{Cat.}    & PSNR     & 17.33     & 18.30 & 16.87 & 17.13    & 17.83   & 23.39  & 19.62  & 19.05  & 19.73   & 21.02 \\ 
                           &                          & SSIM     & 0.502     & 0.554 & 0.525 & 0.491    & 0.498   & 0.579  & 0.485  & 0.606  & 0.584   & 0.684 \\ 
                           &                          & LPIPS    & 0.449     & 0.415 & 0.446 & 0.394    & 0.413   & 0.427  & 0.420  & 0.376  & 0.443   & 0.371 \\ 
                           &                          &    $\mathcal{L}_{1}^{\text{depth}}$  & 0.417     & 0.394 & 0.392 & 0.169    & 0.096   & 0.234  & 0.177  & 0.352  & 0.336   & 0.331 \\ \cline{2-13} 
                           & \multirow{4}{*}{All*-ft} & PSNR     & 22.12     & 27.53 & 26.28 & 25.80    & 22.89   & 30.03  & 26.33  & 29.15  & 26.74   & 32.00 \\ 
                           &                          & SSIM     & 0.683     & 0.829 & 0.769 & 0.834    & 0.686   & 0.814  & 0.764  & 0.845  & 0.815   & 0.885 \\ 
                           &                          & LPIPS    & 0.298     & 0.177 & 0.238 & 0.152    & 0.267   & 0.211  & 0.199  & 0.177  & 0.268   & 0.163 \\ 
                           &                          &    $\mathcal{L}_{1}^{\text{depth}}$  & 0.232     & 0.051 & 0.079 & 0.083    & 0.054   & 0.036  & 0.075  & 0.051  & 0.073   & 0.080 \\ \cline{2-13} 
                           & \multirow{4}{*}{Cat.-ft} & PSNR     & 21.90     & 26.47 & 24.83 & 22.46    & 24.74   & 27.68  & 26.41  & 25.37  & 26.61   & 30.18 \\ 
                           &                          & SSIM     & 0.678     & 0.804 & 0.739 & 0.707    & 0.755   & 0.727  & 0.766  & 0.745  & 0.813   & 0.861 \\ 
                           &                          & LPIPS    & 0.301     & 0.195 & 0.261 & 0.233    & 0.210   & 0.280  & 0.197  & 0.254  & 0.266   & 0.184 \\ 
                           &                          &    $\mathcal{L}_{1}^{\text{depth}}$  & 0.225     & 0.049 & 0.070 & 0.101    & 0.046   & 0.063  & 0.062  & 0.195  & 0.065   & 0.111 \\ \midrule
\multirow{8}{*}{pixelNeRF~\cite{AlexYu2021pixelNeRFNR}} & \multirow{4}{*}{All*}    & PSNR     & 19.77     & 21.54 & 20.77 & 20.15    & 20.93   & 24.73  & 21.78  & 23.48  & 21.30   & 27.18 \\ 
                           &                          & SSIM     & 0.647     & 0.701 & 0.690 & 0.661    & 0.671   & 0.666  & 0.606  & 0.748  & 0.696   & 0.833 \\ 
                           &                          & LPIPS    & 0.377     & 0.331 & 0.363 & 0.315    & 0.339   & 0.393  & 0.370  & 0.283  & 0.381   & 0.269 \\ 
                           &                          &    $\mathcal{L}_{1}^{\text{depth}}$  & 0.142     & 0.131 & 0.141 & 0.109    & 0.073   & 0.085  & 0.114  & 0.065  & 0.175   & 0.061 \\ \cline{2-13} 
                           & \multirow{4}{*}{Cat.}    & PSNR     & 19.91     & 20.93 & 17.55 & 20.20    & 19.63   & 24.16  & 20.80  & 18.59  & 19.84   & 24.96 \\ 
                           &                          & SSIM     & 0.685     & 0.702 & 0.622 & 0.686    & 0.645   & 0.662  & 0.606  & 0.667  & 0.657   & 0.828 \\ 
                           &                          & LPIPS    & 0.332     & 0.330 & 0.426 & 0.275    & 0.348   & 0.392  & 0.367  & 0.342  & 0.420   & 0.249 \\ 
                           &                          &    $\mathcal{L}_{1}^{\text{depth}}$  & 0.136     & 0.224 & 0.364 & 0.119    & 0.142   & 0.152  & 0.243  & 0.181  & 0.336   & 0.054 \\ 
\bottomrule
\end{tabular}
}
  \label{tab:sparse_nerf_full}
  \vspace{-10pt}
\end{table*}


\definecolor{ultramarine}{rgb}{0.07, 0.04, 0.56} 
\definecolor{battleshipgrey}{rgb}{0.52, 0.52, 0.51}

\newcommand{\increase}{\textcolor{ultramarine}}
\newcommand{\decrease}{\textcolor{battleshipgrey}}

\begin{table*}[t]
\centering\scriptsize
\caption{\textbf{Unaligned Cross-scene novel view synthesis results of pixelNeRF-U~\cite{AlexYu2021pixelNeRFNR} on 10 categories.} }
  \resizebox{.98\textwidth}{!} {
\begin{tabular}{c|c|cccccccccc}
\toprule
 Train  & Metric   & toy train & bread & cake  & toy boat & hot dog & wallet & pitaya & squash & handbag & apple \\ \midrule
\multirow{6}{*}{All*} & \multirow{2}{*}{PSNR} & 18.81  & 19.92 & 19.86 & 19.54  & 19.64   & 20.31  & 20.44 & 20.74  & 20.79  & 21.21 \\ 
   &   & \decrease{\scriptsize{-0.96}}  & \decrease{\scriptsize{-1.62}} & \decrease{\scriptsize{-0.91}} & \decrease{\scriptsize{-0.29}}  & \decrease{\scriptsize{-1.29}}   & \decrease{\scriptsize{-4.42}}  & \decrease{\scriptsize{-1.34}} & \decrease{\scriptsize{-2.74}}  & \decrease{\scriptsize{-0.51}}   & \decrease{\scriptsize{-5.97}} \\ 
   & \multirow{2}{*}{SSIM}  & 0.591 & 0.625 & 0.636 & 0.626    & 0.627   & 0.628  & 0.619  & 0.631  & 0.635   & 0.650 \\ 
   &   & \decrease{\scriptsize{-0.056}}  & \decrease{\scriptsize{-0.076}} & \decrease{\scriptsize{-0.054}} & \decrease{\scriptsize{-0.035}}  & \decrease{\scriptsize{-0.044}}   & \decrease{\scriptsize{-0.038}}  & \increase{\scriptsize{+0.013}} & \decrease{\scriptsize{-0.117}}  & \decrease{\scriptsize{-0.061}}   & \decrease{\scriptsize{-0.183}} \\ 
   & \multirow{2}{*}{LPIPS}    & 0.432 & 0.406 & 0.405 & 0.398    & 0.397   & 0.401  & 0.405  & 0.394  & 0.397   & 0.390 \\ 
   &   & \decrease{\scriptsize{-0.055}}  & \decrease{\scriptsize{-0.075}} & \decrease{\scriptsize{-0.042}} & \decrease{\scriptsize{-0.083}}  & \decrease{\scriptsize{-0.058}}   & \decrease{\scriptsize{-0.008}}  & \decrease{\scriptsize{-0.035}} & \decrease{\scriptsize{-0.111}}  & \decrease{\scriptsize{-0.016}}   & \decrease{\scriptsize{-0.121}} \\ 
    &  \multirow{2}{*}{    $\mathcal{L}_{1}^{\text{depth}}$}  & 0.145     & 0.118 & 0.123 & 0.132    & 0.122   & 0.120  & 0.119  & 0.113  & 0.121   & 0.117 \\ 
&   & \decrease{\scriptsize{-0.003}}  & \increase{\scriptsize{+0.013}} & \increase{\scriptsize{+0.018}} & \decrease{\scriptsize{-0.023}}  & \decrease{\scriptsize{-0.049}}  & \decrease{\scriptsize{-0.035}} & \decrease{\scriptsize{-0.005}} & \decrease{\scriptsize{-0.048}}  & \increase{\scriptsize{+0.054}}   & \decrease{\scriptsize{-0.056}} \\ 

\midrule
    
 \multirow{6}{*}{Cat.}    & \multirow{2}{*}{PSNR}     & 19.36 & 19.03 & 18.46 & 18.45    & 18.53   & 19.41  & 19.51  & 19.34  & 19.38   & 19.58 \\ 
  &   & \decrease{\scriptsize{-0.55}}  & \decrease{\scriptsize{-1.90}} & \decrease{\scriptsize{-0.91}} & \decrease{\scriptsize{-1.75}}  & \decrease{\scriptsize{-1.10}}   & \decrease{\scriptsize{-4.75}}  & \decrease{\scriptsize{-1.29}} & \decrease{\scriptsize{-0.75}}  & \decrease{\scriptsize{-0.46}}   & \decrease{\scriptsize{-5.38}} \\ 
   & \multirow{2}{*}{SSIM}     & 0.637     & 0.636 & 0.626 & 0.624    & 0.623   & 0.625  & 0.616  & 0.614  & 0.618   & 0.631 \\ 
   &   & \decrease{\scriptsize{-0.048}}  & \decrease{\scriptsize{-0.066}} & \increase{\scriptsize{+0.004}} & \decrease{\scriptsize{-0.062}}  & \decrease{\scriptsize{-0.022}}   & \decrease{\scriptsize{-0.037}}  & \increase{\scriptsize{+0.010}} & \decrease{\scriptsize{-0.053}}  & \decrease{\scriptsize{-0.039}}   & \decrease{\scriptsize{-0.197}} \\ 
 & \multirow{2}{*}{LPIPS}    & 0.392     & 0.402 & 0.415 & 0.400    & 0.396   & 0.399  & 0.403  & 0.404  & 0.408   & 0.404 \\ 
    &   & \decrease{\scriptsize{-0.060}}  & \decrease{\scriptsize{-0.072}} & \increase{\scriptsize{+0.011}} & \decrease{\scriptsize{-0.125}}  & \decrease{\scriptsize{-0.048}}   & \decrease{\scriptsize{-0.007}}  & \decrease{\scriptsize{-0.036}} & \decrease{\scriptsize{-0.062}}  & \increase{\scriptsize{+0.012}}   & \decrease{\scriptsize{-0.155}} \\ 
   &  \multirow{2}{*}{  $\mathcal{L}_{1}^{\text{depth}}$}  & 0.172     & 0.219 & 0.260 & 0.262    & 0.247   & 0.251  & 0.252  & 0.286  & 0.293  & 0.276 \\ 
&   & \decrease{\scriptsize{-0.036}}  & \increase{\scriptsize{+0.005}} & \increase{\scriptsize{+0.104}} & \decrease{\scriptsize{-0.143}}  & \decrease{\scriptsize{-0.105}}   & \decrease{\scriptsize{-0.099}}  & \decrease{\scriptsize{-0.009}} & \decrease{\scriptsize{-0.105}}  & \increase{\scriptsize{+0.043}}   & \decrease{\scriptsize{-0.222}} \\ 
\bottomrule
\end{tabular}
}
  \label{tab:sparse_nerf_unaligned_full}%
\end{table*}


\begin{figure*}[t]
    \centering
    \includegraphics[width=1.\linewidth]{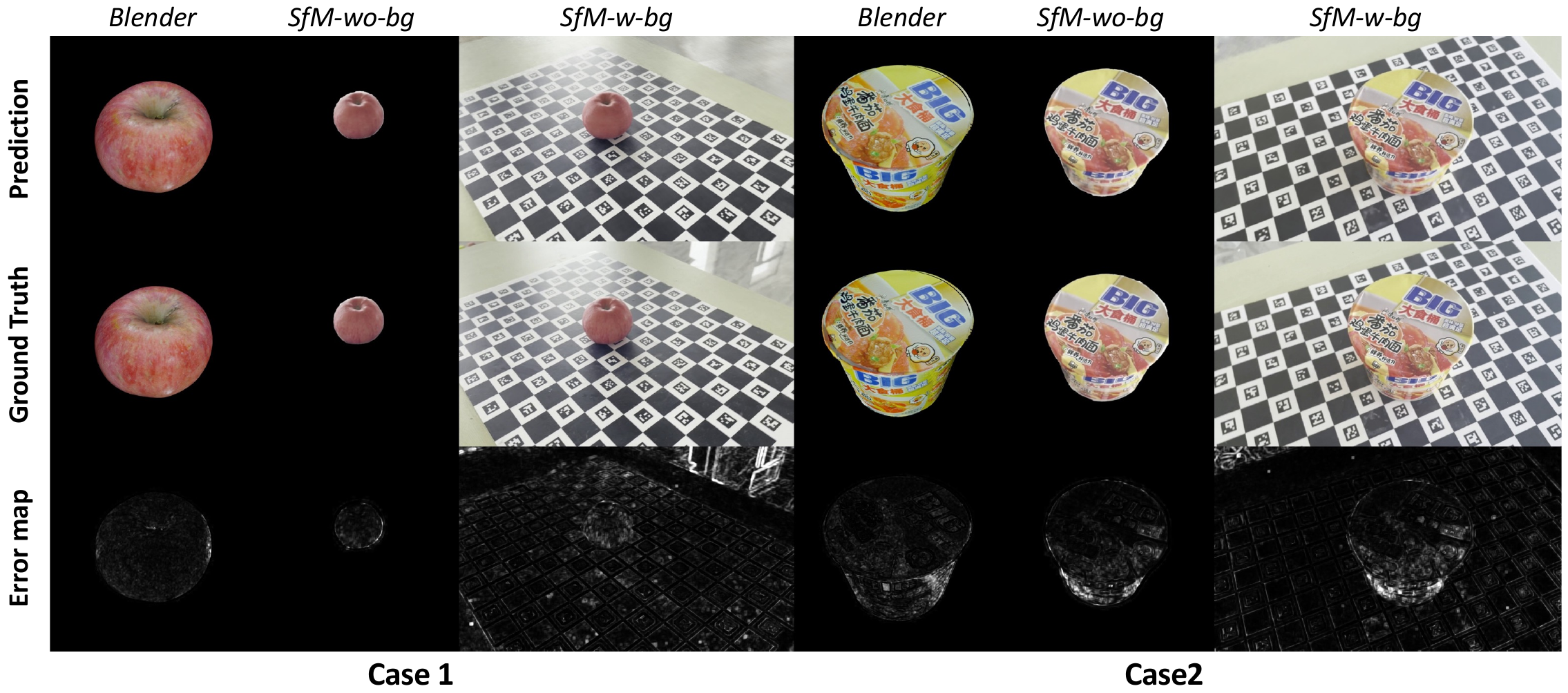}
    \setlength{\abovecaptionskip}{0mm}
    \caption{\small
    \textbf{Qualitative comparisons of NVS on the same scenes with different data dypes.}
    }
    \label{fig:single-scene NVS supp}
    \vspace{-10pt}
\end{figure*}

\begin{figure*}[t]
    \centering
    \includegraphics[width=1.\linewidth]{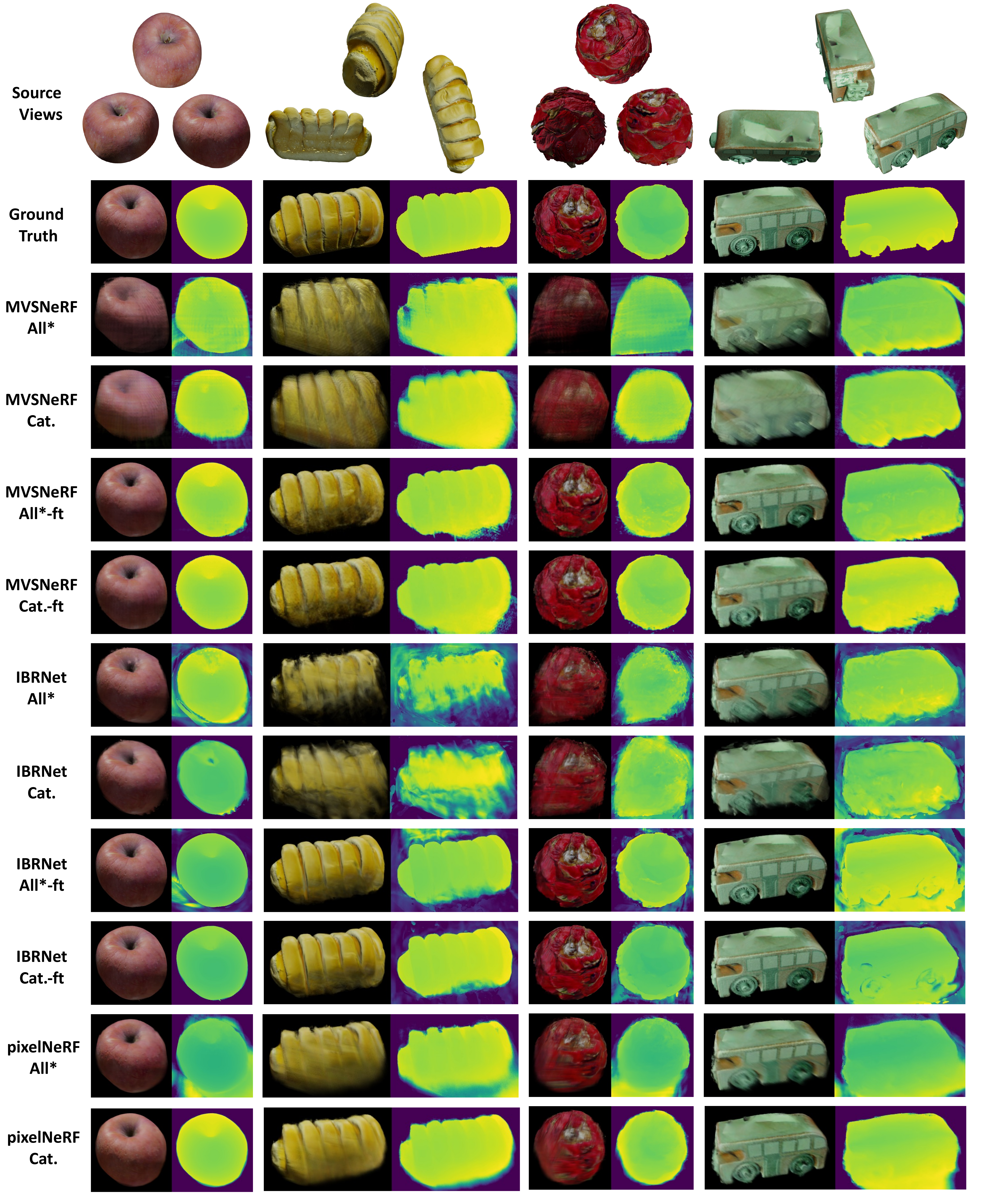}
    \setlength{\abovecaptionskip}{0mm}
    \caption{\small
    \textbf{Qualitative comparisons of several cross-scene NVS methods in different scenes from our dataset.}
    }
    \label{fig:sparse_nvs}
    \vspace{-10pt}
\end{figure*}

\begin{figure}[t]
    \centering
    \includegraphics[width=1\linewidth]{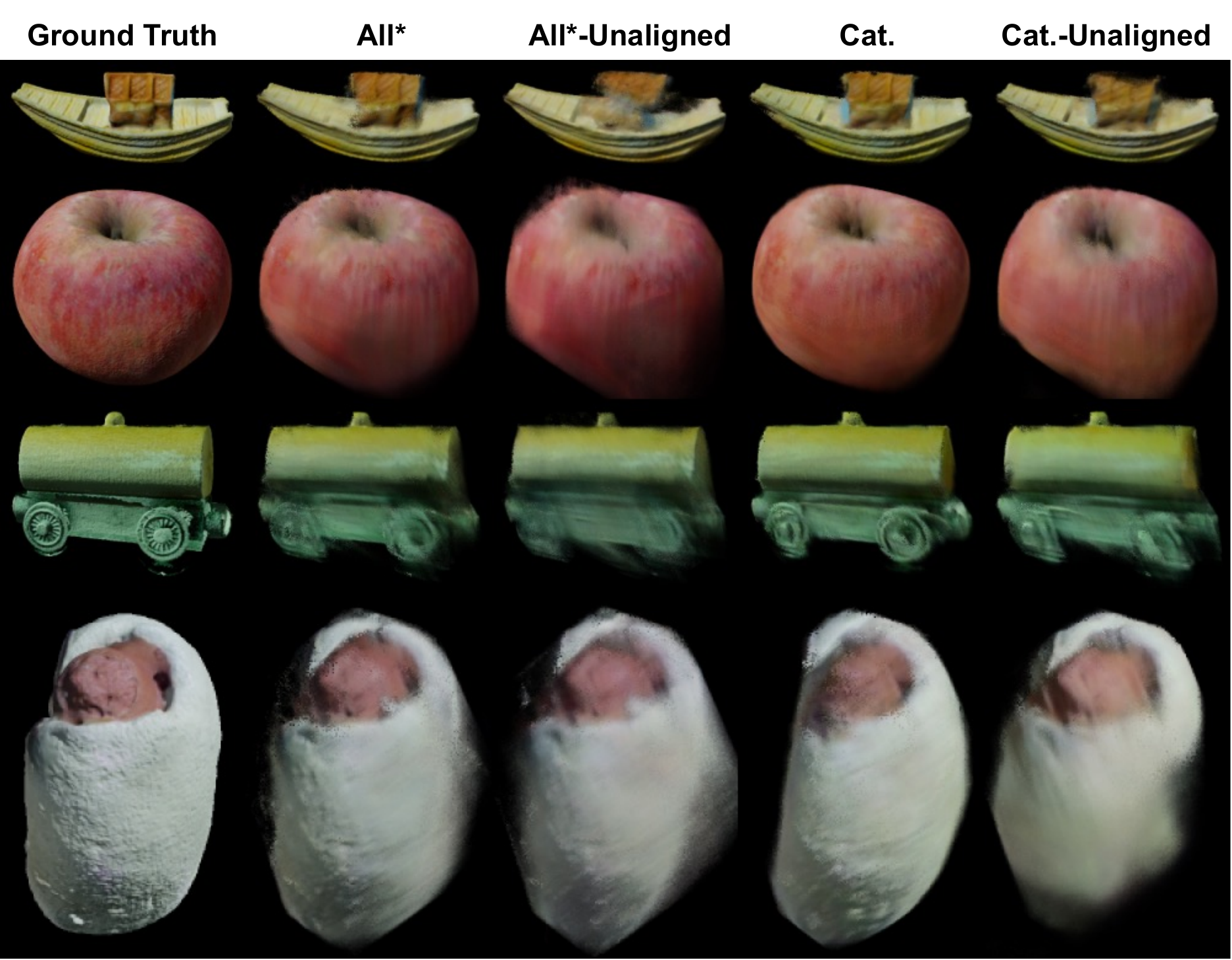}
    \setlength{\abovecaptionskip}{0mm}
    \caption{\small
    \textbf{Qualitative comparison of pixelNeRF-U and pixelNeRF.} The former shows a more blurry and irregular-shaped appearance.
    }
    \label{fig:pixelNeRF-U}
    \vspace{-10pt}
\end{figure}

\noindent{\textbf{Implementation Details.}} 
We use the official code and default settings by NeRF~\cite{mildenhall2020nerf}, Mip-NeRF~\cite{JonathanTBarron2021MipNeRF3U}, and Plenoxels~\cite{yu2021plenoxels} in this section. For NeRF, we re-weight the foreground and background contents by 1:0.5 to avoid all-black output. For Plenoxel on the SfM data with background, we enable the background model provided by the official code to model the background area.

\noindent{\textbf{Qualitative Comparisons of NVS on rendered images.}}
We describe the performance of three representative methods in the main text, and we provide some qualitative comparisons here in Figure~ \ref{fig:single_nvs}, accordingly. Plenoxels are especially good at modelling high-frequency textures (\eg, the coconut), while it is less robust then NeRF and mip-NeRF when dealing with dark textures and concave geometry, suffering from inaccurate geometry. Our dataset helps to provide a comprehensive evaluation of different methods.

\noindent{\textbf{Comparisons of NVS on rendered images and iPhone videos.}} We conduct qualitative and quantitative evaluations on novel view synthesis with several scenes under different data types, including \emph{SfM-wo-bg}, \emph{SfM-w-bg} and \emph{Blender}. The \emph{SfM-wo-bg} and \emph{SfM-w-bg} settings use images sampled from iPhone videos and camera parameters generated by COLMAP. The difference between them is whether the background is included. The \emph{Blender} data are rendered by Blender~\cite{blender}.
Since the image resolutions and foreground proportions are different among the data types, we calculate the PSNR metric only in the foreground area for \emph{SfM-wo-bg} data and \emph{Blender} data, whereas for \emph{SfM-w-bg} data, every pixel in the image is included PSNR calculation.

Based on the qualitative comparisons in Figure~\ref{fig:single-scene NVS supp}, we observe that for both two selected scenes, the predicted novel view image under the \emph{Blender} setting achieves the best visual quality, resulting in the highest PSNR in Table~\ref{tab:single_nerf_scenario}. When comparing the two SfM based data types, we find that the quality of the foreground object from the \emph{SfM-wo-bg} data is only slightly better than the other, while the high background error under the \emph{SfM-w-bg} setting leads to a significant drop in performance, as shown in Table~\ref{tab:single_nerf_scenario}. The experimental results shed light on how real-captured videos introduce further challenges to NeRF-like methods. We demonstrate that performing robust novel view synthesis with casually captured videos will be an important and practical topic.

\begin{figure*}[t]
    \centering
    \includegraphics[width=0.9\linewidth]{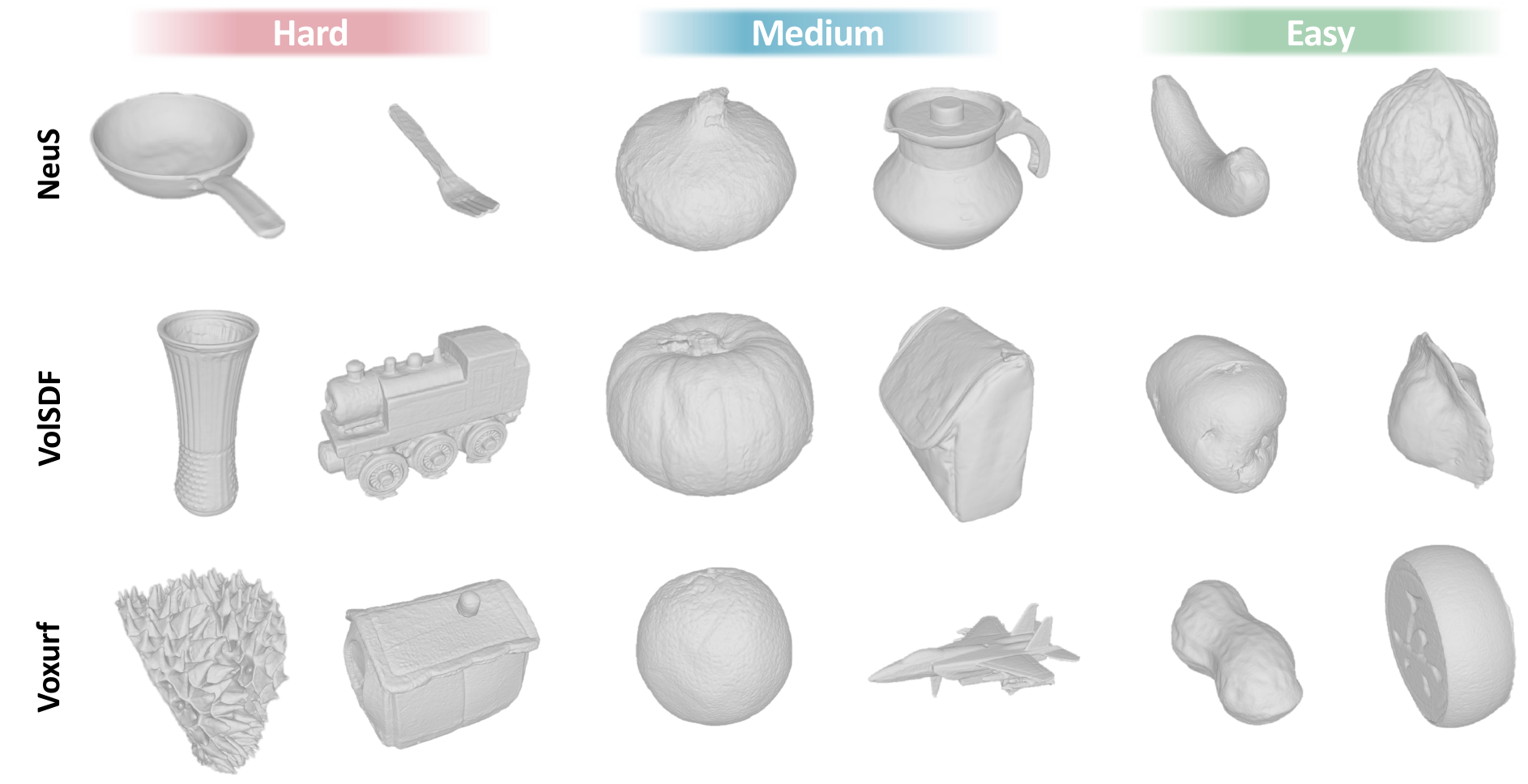}
    \setlength{\abovecaptionskip}{0mm}
    \caption{\small
    \textbf{Examples from different difficulty levels in surface reconstruction.}
    }
    \label{fig:hardness_examples}
    \vspace{-10pt}
\end{figure*}

\subsubsection{Cross-Scene NVS}

\noindent{\textbf{Implementation Details.}} 
 We use the official codes to evaluate three benchmarks on 10 categories, \ie, toy train, bread, cake, toy boat, hot dog, wallet, pitaya, squash, handbag, and apple. We split three scenes from each category as a test-set, and the remaining scenes are used as a train-set. During training, we randomly sample rays from scenes in the train-set of each category and use Adam~\cite{kingma2014adam} optimizer. For a fair comparison, we evaluate these methods with the same source views, \ie, 3 views from nearby 30 views (explained in Sec.~C.3.2) by FPS sampling. Then in a scene with 100 rendered views, we exclude these 3 source views and select 10 test views from the remaining 97 views by FPS criteria again. For MVSNeRF, we pretrain the `All*' with total 300k iterations, and the `Cat.' with 20k to 40k iterations depending on the number of scenes. In finetuning stage, we take 3 views as input and additional 13 views sampling for per-scene optimization. Each scene is finetuned for 15k iterations. For IBRNet, we pretrain the `All*' with 300k iterations, and the `Cat.' with 50k iterations. After cross-scene training, we further finetune the model with 15k iterations on each test scene. For pixelNeRF, we train the `All*' with 400k iterations, and the `Cat.' with 12k to 30k iterations depending on the number of scenes. All methods sample rays within a tight foreground bounding box around the object. 

\noindent{\textbf{Detailed Comparisons.}} The full evaluation results are presented in Table~\ref{tab:sparse_nerf_full}.  
We additionally provide qualitative comparisons of 4 cases, each with rendered RGB and depth, as shown in Figure~\ref{fig:sparse_nvs} (we leave an extra 15 pixels of each edge). We evaluate depth within the foreground masks. From the visualization, it may seem that methods w/ 'Cat.' generate more accurate contour than that w/ `All*', contradicting the statement that methods w/ `All*' can learn a better geometric cue than that w/ `Cat.' in the main context. However, we find that within the masks, the depth of the former is generally more precise than that of the latter, obviously illustrated by ``pitaya'' (the third case) in pixelNeRF. It may raise an interesting research topic of how generic methods can perform both accurately in shape contour and geometry. After slightly finetuning MVSNeRF and IBRNet on a test scene, these methods achieve comparable performance with scene-specific methods, \eg, NeRF.

\noindent{\textbf{Results on Unaligned Coordinate System.}} We additionally provide a more challenging setting by evaluating Cross-Scene NVS on an unaligned coordinate system rather than in a perfectly predefined canonical space. Specifically, we examine pixelNeRF-U~\cite{AlexYu2021pixelNeRFNR}, where the coordinate system of each object is randomly rotated by $\theta \left( \sim 60^{\circ} \cdot \mathcal N \left( 0, 1\right) \right)$ in three axes and translated by $\left[ 0.5, 0.5, 0.5 \right] \cdot \mathcal N \left( 0, 1\right)$. As detailedly illustrated in Table~\ref{tab:sparse_nerf_unaligned_full} and Figure~\ref{fig:pixelNeRF-U}, the PSNR drops with All*:22.16$\rightarrow$21.20, Cat.: 20.65$\rightarrow$19.58, particularly for apple and wallet, and the geometry also suffers except for bread, cake, and handbag, resulting in a gennerally more blurry and irregular-shaped appearance. We infer that since xyz is fed into the network, the coordinates will implicitly store category-specific priors, \eg, a specific sampled 3D location in canonical space will learn the prior of head or tail (other elements) of toy train. Thus the misalignment will tend to impair this learned variance of the rigid scene. In our experiment, we manually perform non-alignment in a regular mathematical manner, we believe this impairment will become more severe when applied to a naturally-unaligned coordinate system.
\subsection{Neural Surface Reconstruction}
\subsubsection{Dense-View Surface Reconstruction}
\begin{figure*}[t]
    \centering
    \includegraphics[width=0.9\linewidth]{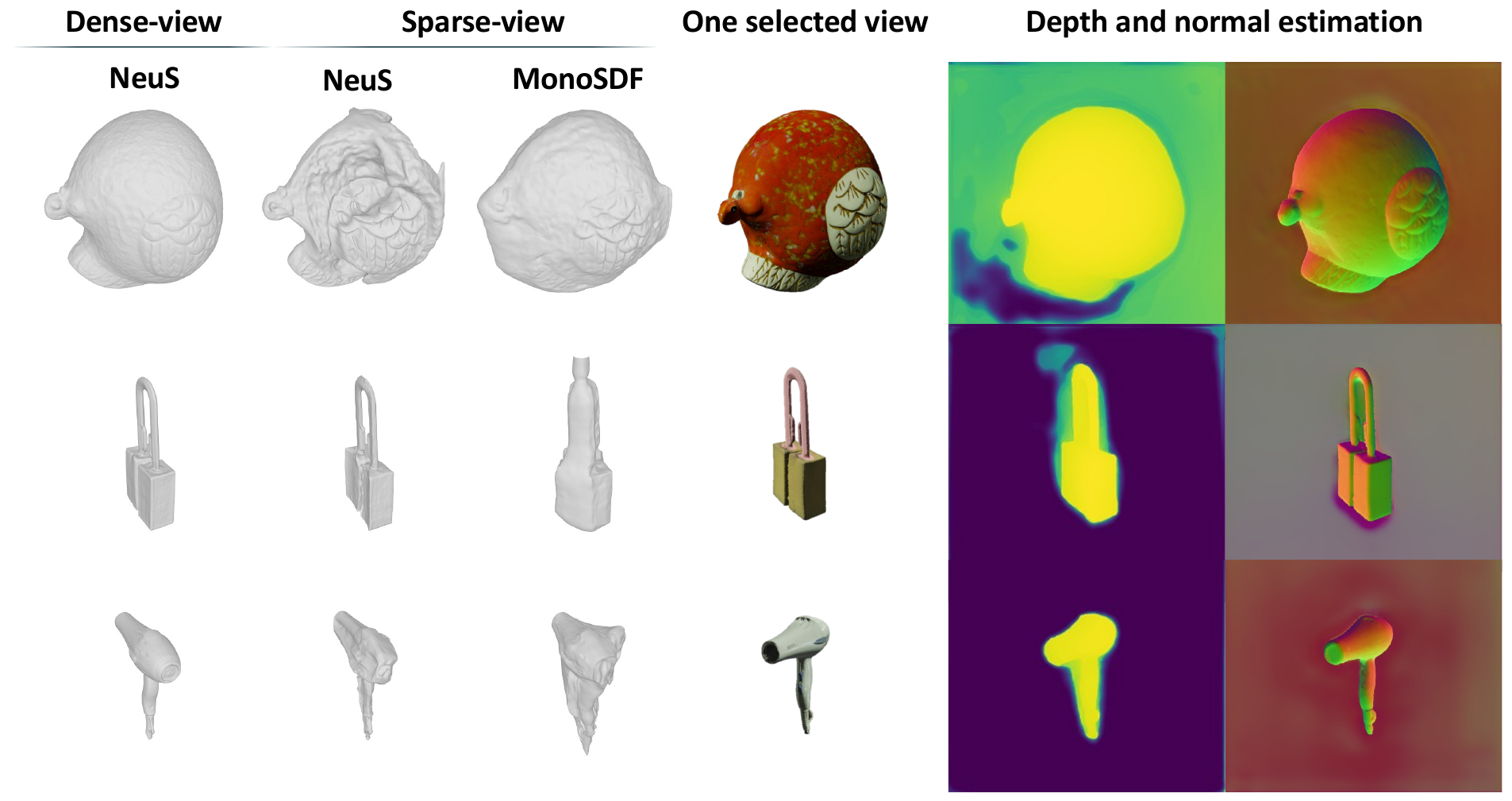}
    \setlength{\abovecaptionskip}{0mm}
    \caption{\small
    \textbf{A comparison of sparse-view surface reconstruction between NeuS and MonoSDF.} The estimated depth and normal maps used by MonoSDF are shown on the right.
    }
    \label{fig:mono_depth-estimation}
    \vspace{-10pt}
\end{figure*}

\begin{figure}[t]
    \centering
    \includegraphics[width=1.\linewidth]{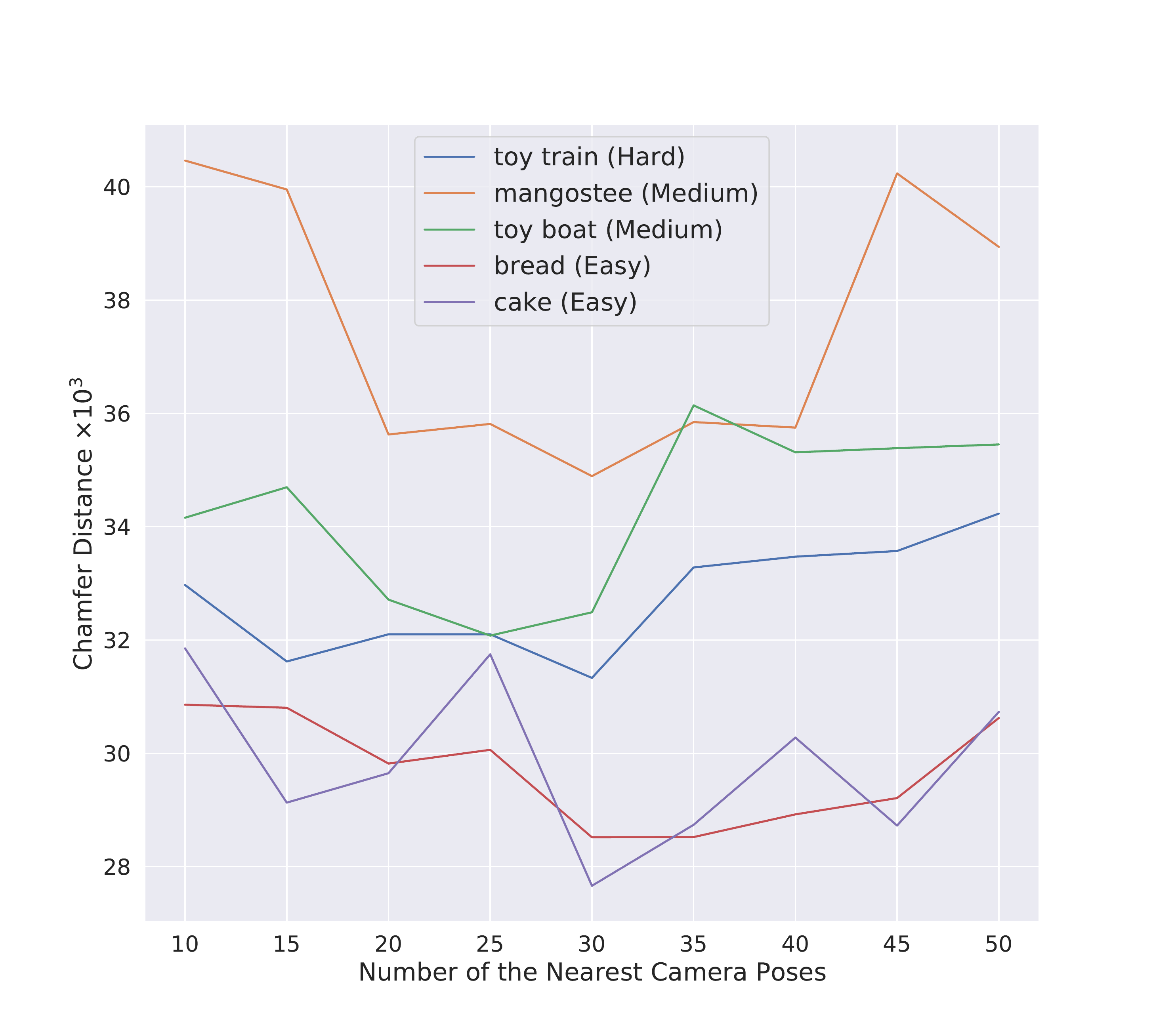}
    \setlength{\abovecaptionskip}{0mm}
    \caption{\small
    \textbf{Geometric quality with regard to view selection range.}
    }
    \label{fig:view_selection_range}
    \vspace{-10pt}
\end{figure}

\noindent{\textbf{Implementation Details.}} 
We use the publicly available code for NeuS~\cite{wang2021neus} and VolSDF~\cite{yariv2021volume}, and we use the code provided by the authors for Voxurf~\cite{wu2022voxurf}, training with for a standard number of iteration on each of them. For all the methods, we do not involve the mask loss as supervision. Each scene is trained on 100 views.
We use the Chamfer Distance between the reconstructed surface and the ground truth mesh for evaluation. The distance is calculated in a normalized space (all coordinates lying within $[-1, 1]$). We clip the distance by 0.1 to alleviate the huge effect of outliers. We will release the standard evaluation code.

\noindent{\textbf{Qualitative Comparisons.}}
In the main text, we split the categories into three difficulty levels, namely \textit{hard}, \textit{medium}, and \textit{easy}. Figure~\ref{fig:hardness_examples} shows some examples from each level. We observe that the ``hard'' examples usually suffer from dark and low-texture appearance (\eg, the pan), concave geometry (\eg, the vase and the kennel), and complex or thin structures (\eg, the durian, the fork, and the toy train). The ``medium'' and ``easy'' cases usually have a simple geometry with proper texture. The wide exploration of geometry and textures of the dataset helps to provide a comprehensive evaluation of different methods.

\subsubsection{Sparse-View Surface Reconstruction}
\noindent{\textbf{Implementation Details.}} 
For NeuS~\cite{wang2021neus} and MonoSDF~\cite{ZehaoYu2022MonoSDFEM}, we use FPS sampling to sample 3 views from all the 100 views. We train 10k iterations for NeuS and 500 epochs for MonoSDF, both being reduced from the original setting due to the few-view input. 
For SparseNeuS~\cite{XiaoxiaoLong2022SparseNeuSFG}, we fix the first three examples in each category as the testing set and skip them when training. We conduct FPS among the nearest 30 camera poses from a random reference view at inference time. The fine-tuning stage of SparseNeuS is not stable: the training usually collapses before convergence, and the issue also exists for the officially used DTU dataset. So we report the results via direct inference for all the experiments.

\noindent{\textbf{Detailed Comparisons.}}
In Table 6 of the main text, we surprisingly find that NeuS can serve as a strong baseline under the sparse-view setting without bells and whistles. MonoSDF is enhanced by depth and normal estimations from pre-trained networks~\cite{eftekhar2021omnidata}, and it claims a superior performance on DTU with only 3 views as input. However, MonoSDF does not seem to perform as well as NeuS in OmniObject3D. 

As shown in Figure~\ref{fig:mono_depth-estimation}, the NeuS baseline with FPS sampling is especially good at dealing with thin structures: the wide-spread views together with the black backgrounds help to bound the geometry well. However, the depth estimation is especially inaccurate in these scenarios, which is probably caused by the gap between the training and testing images of the depth estimation network. Nevertheless, it shows great performance in some cases for maintaining a coherent shape and adding some geometry details. It is an interesting problem to explore a robust usage of the estimated geometry cues under different circumstances.

\begin{table}[t]
  \centering
  \caption{\textbf{Sparse-view surface reconstruction results with a range of views.}}
    \begin{tabular}{l|cccc}
    \toprule
    \multicolumn{1}{c|}{\multirow{2}[4]{*}{Method}} & \multicolumn{4}{c}{Chamfer Distance $\times 10^3$ ($\downarrow$)} \\
\cmidrule{2-5}          & 2 views & 3 views & 5 views & 8 views \\
    \midrule
    \multicolumn{1}{c|}{NeuS~\cite{wang2021neus}} & 41.06 & 27.3  & 12.65 & 7.96 \\
    MonoSDF~\cite{ZehaoYu2022MonoSDFEM} & 45.35 & 34.68 & 23.02 & 18.97 \\
    \bottomrule
    \end{tabular}%
  \vspace{-10pt}
  \label{tab:different_views}%
\end{table}%

\noindent{\textbf{Sparse-view surface reconstruction with a range of view numbers.}} 
In addition to the default setting of 3 views, we try a range of views (\ie, 2, 3, 5, 8 views) with FPS sampling for NeuS~\cite{wang2021neus} and MonoSDF~\cite{ZehaoYu2022MonoSDFEM}, and the results are shown in Table.~\ref{tab:different_views}. For NeuS, we observe a significant improvement in accuracy as the view number increases from 2 to 8, but the 8-view setting (7.96) is still worse than the 100-view setting (6.09) with a clear margin. For MonoSDF, the improvement begins to slow down when lifting from 5 views to 8 views. This problem is probably due to the inaccurate depth guidance, as described above.

\noindent{\textbf{View Selection Range for Cost Volume Initialization.}} In MVSNeRF~\cite{AnpeiChen2021MVSNeRFFG}, due to occlusions, initialized local cost volume feature is inconsistent with large viewpoint changes, causing poor geometry extracted from the global density field. One na\"ive solution is to decrease the interval distance between source views. Although the constructed local feature will accordingly be more consistent as the occlusion region reduces, it will encode less source context. To make a trade-off between feature consistency and richness of encoded information, we conduct a comparison on how the extracted mesh will perform with the number of the nearest source views in FPS on 15 random categories from three levels of ``difficulty''.  We filter the categories with averaged CD $\geq$ 0.04, whose geometries are too poor to rely on. Finally, we remain 5 classes as shown in Figure~\ref{fig:view_selection_range}. The geometric quality shows a fluctuating trend of decreasing and then increasing with regard to the view range. As a result, we pick up ``30'' as a proper view selection range. Similarly, we find that ``30'' can also be applied to SparseNeuS~\cite{XiaoxiaoLong2022SparseNeuSFG} for cascaded geometry volume construction.

\subsection{3D Object Generation}
\noindent{\textbf{Implementation Details.}}
We use the official code by GET3D~\cite{gao2022get3d} to train all the models. We prepare the multi-view image dataset by rendering 24 inward-facing multi-view images per object with Blender~\cite{blender}. For the large subset with 100 categories, we train 7k iterations with MSE loss and Adam optimizer; we train 3k iterations on smaller subsets (\eg, \textit{furniture}, \textit{fruits}, and \textit{toys}). 

\begin{figure}[t]
    \centering
    \includegraphics[width=0.8\linewidth]{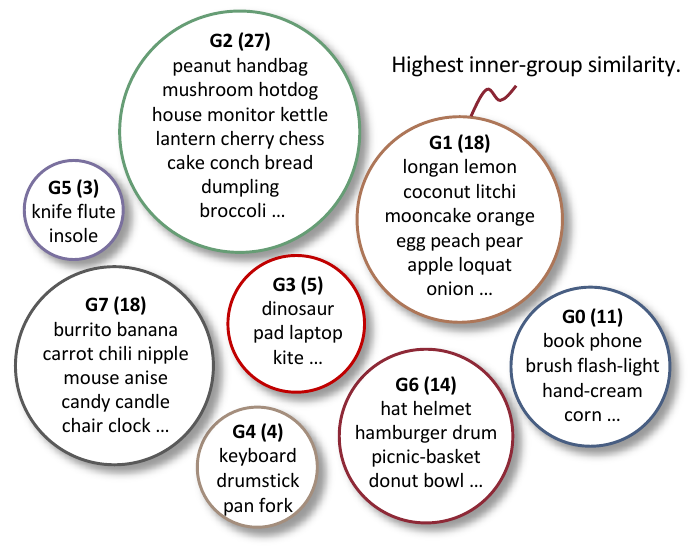}
    \setlength{\abovecaptionskip}{0mm}
    \caption{\small
    \textbf{Categories in each group after the KMeans clustering.} Categories in Group 1 are highly similar to each other, while those in Group 2 bear a high inner-group divergence. 
    }
    \label{fig:group_names}
    \vspace{-10pt}
\end{figure}

\begin{figure}[t]
    \centering
    \includegraphics[width=1.\linewidth]{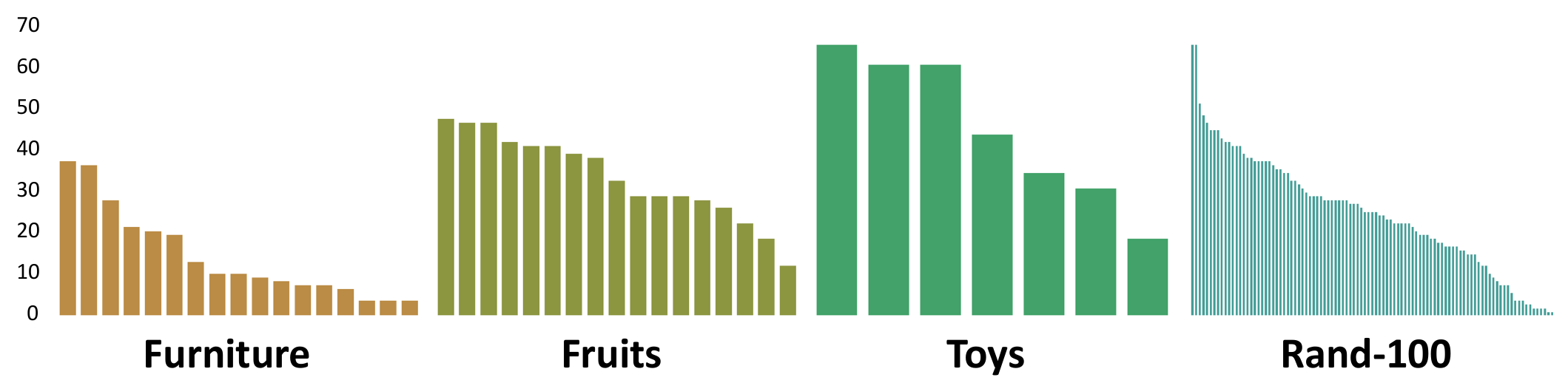}
    \setlength{\abovecaptionskip}{0mm}
    \caption{\small
    \textbf{Distributions of the four subsets.}
    }
    \label{fig:4_group_distributions}
    \vspace{-10pt}
\end{figure}

\noindent{\textbf{Additional Experimental Results and Discussions.}}
We study the semantic distribution in the main text, where we use KMeans to cluster 100 random categories into 8 groups, as shown in Figure~\ref{fig:group_names}. We can observe that Group 2 has the largest number of categories, while they suffer from a high inner-group divergence (\eg, the peanut, handbag, mushroom, and hot dog). In contrast, Group 1 contains many fruits, vegetables, and some other categories that are similar in shape. The high inner-group similarity enables them to enhance the learning of each other, and Group 1 is finally able to dominate the generation distribution. The Group-level analysis reveals how cross-class relationships affect the generation distribution, which is a critical factor for generative models trained with large vocabulary datasets like OmniObject3D.
We also provide the distribution of the four subsets used in this section in Figure~\ref{fig:4_group_distributions}.

\begin{figure}[t]
    \centering
    \includegraphics[width=1.0\linewidth]{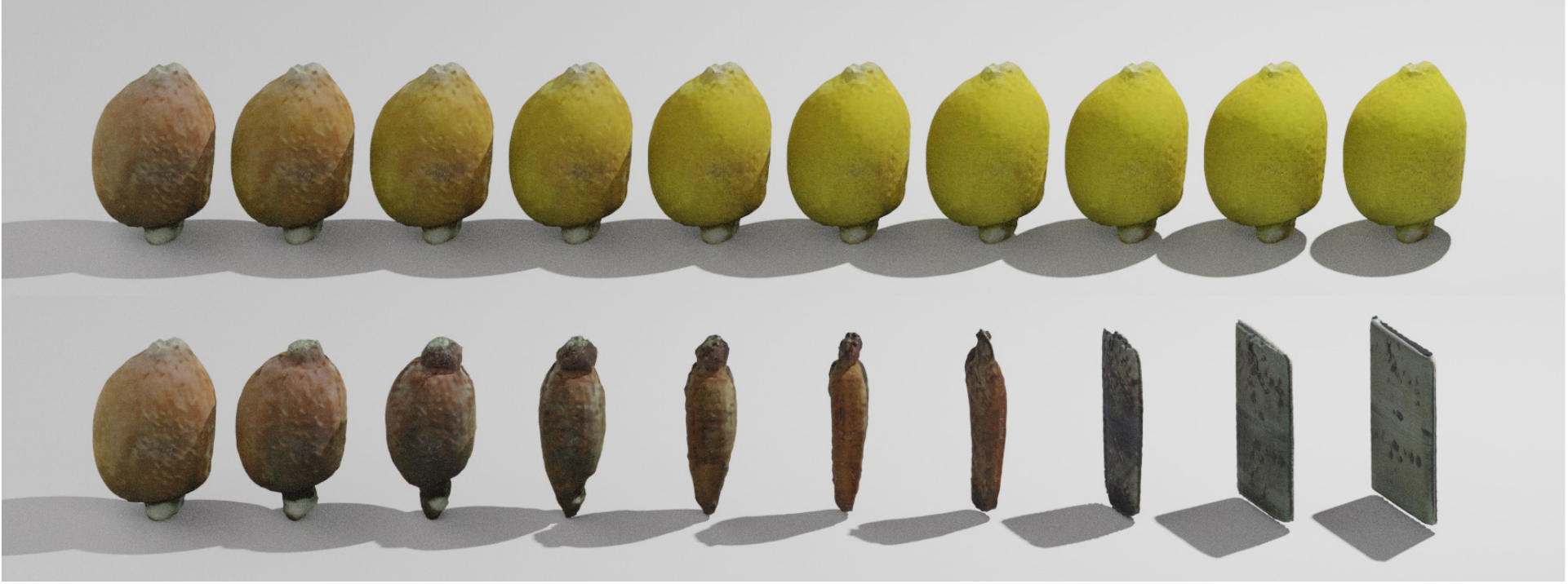}
    \setlength{\abovecaptionskip}{0mm}
    \caption{\small
    \textbf{Shape Interpolation.} In the first row, we keep the latent code of geometry fixed and interpolate the latent code of texture; in the second row, we keep the latent code of texture fixed and interpolate the latent code of geometry.
    }
    \label{fig:texture_geo_interpolation}
    \vspace{-10pt}
\end{figure}

Finally, we provide disentangled interpolation results in Figure~\ref{fig:texture_geo_interpolation} with geometry latent code and texture latent code, respectively. In the first row, the texture changes with a fixed shape, and the semantic changes accordingly. In the second row, when the geometry changes, the texture is fixed at first while encountering a substantial change along with the geometry at the end. This indicates that the two factors are not fully disentangled, and the geometry code can sometimes affect the texture since the category, geometry, and texture are highly correlated with each other in the dataset.
Meanwhile, we observe that complex textures (\eg, the cover of a book) usually fail to be well generated, which is another challenging problem to be explored in the future.

\newpage

{\small
\bibliographystyle{ieee_fullname}
\bibliography{egbib}

\begin{thebibliography}{100}\itemsep=-1pt

\bibitem{aanaes2016large}
Henrik Aan{\ae}s, Rasmus~Ramsb{\o}l Jensen, George Vogiatzis, Engin Tola, and
  Anders~Bjorholm Dahl.
\newblock Large-scale data for multiple-view stereopsis.
\newblock {\em International Journal of Computer Vision (IJCV)},
  120(2):153--168, 2016.

\bibitem{PanosAchlioptas2017LearningRA}
Panos Achlioptas, Olga Diamanti, Ioannis Mitliagkas, and Leonidas Guibas.
\newblock Learning representations and generative models for 3d point clouds.
\newblock In {\em Proceedings of the International Conference on Machine
  learning (ICML)}, pages 40--49, 2018.

\bibitem{ahmadyan2021objectron}
Adel Ahmadyan, Liangkai Zhang, Artsiom Ablavatski, Jianing Wei, and Matthias
  Grundmann.
\newblock Objectron: A large scale dataset of object-centric videos in the wild
  with pose annotations.
\newblock In {\em Proceedings of the IEEE/CVF Conference on Computer Vision and
  Pattern Recognition (CVPR)}, pages 7822--7831, 2021.

\bibitem{atzmon2019controlling}
Matan Atzmon, Niv Haim, Lior Yariv, Ofer Israelov, Haggai Maron, and Yaron
  Lipman.
\newblock Controlling neural level sets.
\newblock In {\em Advances in Neural Information Processing Systems (NIPS)},
  volume~32, 2019.

\bibitem{barron2021mip}
Jonathan~T Barron, Ben Mildenhall, Matthew Tancik, Peter Hedman, Ricardo
  Martin-Brualla, and Pratul~P Srinivasan.
\newblock Mip-nerf: A multiscale representation for anti-aliasing neural
  radiance fields.
\newblock In {\em Proceedings of the IEEE/CVF International Conference on
  Computer Vision (ICCV)}, pages 5855--5864, 2021.

\bibitem{JonathanTBarron2021MipNeRF3U}
Jonathan~T Barron, Ben Mildenhall, Dor Verbin, Pratul~P Srinivasan, and Peter
  Hedman.
\newblock Mip-nerf 360: Unbounded anti-aliased neural radiance fields.
\newblock In {\em Proceedings of the IEEE/CVF Conference on Computer Vision and
  Pattern Recognition (CVPR)}, pages 5470--5479, 2022.

\bibitem{EricRChan2022EfficientG3}
Eric~R Chan, Connor~Z Lin, Matthew~A Chan, Koki Nagano, Boxiao Pan, Shalini
  De~Mello, Orazio Gallo, Leonidas~J Guibas, Jonathan Tremblay, Sameh Khamis,
  et~al.
\newblock Efficient geometry-aware 3d generative adversarial networks.
\newblock In {\em Proceedings of the IEEE/CVF Conference on Computer Vision and
  Pattern Recognition (CVPR)}, pages 16123--16133, 2022.

\bibitem{EricRChan2020piGANPI}
Eric~R Chan, Marco Monteiro, Petr Kellnhofer, Jiajun Wu, and Gordon Wetzstein.
\newblock pi-gan: Periodic implicit generative adversarial networks for
  3d-aware image synthesis.
\newblock In {\em Proceedings of the IEEE/CVF Conference on Computer Vision and
  Pattern Recognition (CVPR)}, pages 5799--5809, 2021.

\bibitem{chang2015shapenet}
Angel~X Chang, Thomas Funkhouser, Leonidas Guibas, Pat Hanrahan, Qixing Huang,
  Zimo Li, Silvio Savarese, Manolis Savva, Shuran Song, Hao Su, et~al.
\newblock Shapenet: An information-rich 3d model repository.
\newblock {\em arXiv.org}, 1512.03012, 2015.

\bibitem{chen2022tensorf}
Anpei Chen, Zexiang Xu, Andreas Geiger, Jingyi Yu, and Hao Su.
\newblock Tensorf: Tensorial radiance fields.
\newblock In {\em Proceedings of the European Conference on Computer Vision
  (ECCV)}, 2022.

\bibitem{AnpeiChen2021MVSNeRFFG}
Anpei Chen, Zexiang Xu, Fuqiang Zhao, Xiaoshuai Zhang, Fanbo Xiang, Jingyi Yu,
  and Hao Su.
\newblock Mvsnerf: Fast generalizable radiance field reconstruction from
  multi-view stereo.
\newblock In {\em Proceedings of the IEEE/CVF International Conference on
  Computer Vision (ICCV)}, pages 14124--14133, 2021.

\bibitem{WenzhengChen2019LearningTP}
Wenzheng Chen, Huan Ling, Jun Gao, Edward Smith, Jaakko Lehtinen, Alec
  Jacobson, and Sanja Fidler.
\newblock Learning to predict 3d objects with an interpolation-based
  differentiable renderer.
\newblock In {\em Advances in Neural Information Processing Systems (NIPS)},
  volume~32, 2019.

\bibitem{chen2020pointmixup}
Yunlu Chen, Vincent~Tao Hu, Efstratios Gavves, Thomas Mensink, Pascal Mettes,
  Pengwan Yang, and Cees~GM Snoek.
\newblock Pointmixup: Augmentation for point clouds.
\newblock In {\em Proceedings of the European Conference on Computer Vision
  (ECCV)}, pages 330--345, 2020.

\bibitem{ZhiqinChen2018LearningIF}
Zhiqin Chen and Hao Zhang.
\newblock Learning implicit fields for generative shape modeling.
\newblock In {\em Proceedings of the IEEE/CVF Conference on Computer Vision and
  Pattern Recognition (CVPR)}, pages 5939--5948, 2019.

\bibitem{chen2019learning}
Zhiqin Chen and Hao Zhang.
\newblock Learning implicit fields for generative shape modeling.
\newblock In {\em Proceedings of the IEEE/CVF Conference on Computer Vision and
  Pattern Recognition (CVPR)}, pages 5939--5948, 2019.

\bibitem{collins2022abo}
Jasmine Collins, Shubham Goel, Kenan Deng, Achleshwar Luthra, Leon Xu, Erhan
  Gundogdu, Xi Zhang, Tomas F~Yago Vicente, Thomas Dideriksen, Himanshu Arora,
  et~al.
\newblock Abo: Dataset and benchmarks for real-world 3d object understanding.
\newblock In {\em Proceedings of the IEEE/CVF Conference on Computer Vision and
  Pattern Recognition (CVPR)}, pages 21126--21136, 2022.

\bibitem{blender}
Blender~Online Community.
\newblock Blender - a 3d modelling and rendering package.
\newblock 2018.

\bibitem{francois2021warping}
Fran{\c{c}}ois Darmon, B{\'{e}}n{\'{e}}dicte Bascle, Jean{-}Cl{\'{e}}ment
  Devaux, Pascal Monasse, and Mathieu Aubry.
\newblock Improving neural implicit surfaces geometry with patch warping.
\newblock {\em arXiv.org}, 2112.09648, 2021.

\bibitem{deng2009imagenet}
Jia Deng, Wei Dong, Richard Socher, Li-Jia Li, Kai Li, and Li Fei-Fei.
\newblock Imagenet: A large-scale hierarchical image database.
\newblock In {\em Proceedings of the IEEE/CVF Conference on Computer Vision and
  Pattern Recognition (CVPR)}, pages 248--255, 2009.

\bibitem{PrafullaDhariwal2021DiffusionMB}
Prafulla Dhariwal and Alexander Nichol.
\newblock Diffusion models beat gans on image synthesis.
\newblock In {\em Advances in Neural Information Processing Systems (NIPS)},
  volume~34, pages 8780--8794, 2021.

\bibitem{downs2022google}
Laura Downs, Anthony Francis, Nate Koenig, Brandon Kinman, Ryan Hickman, Krista
  Reymann, Thomas~B McHugh, and Vincent Vanhoucke.
\newblock Google scanned objects: A high-quality dataset of 3d scanned
  household items.
\newblock {\em arXiv.org}, 2204.11918, 2022.

\bibitem{eftekhar2021omnidata}
Ainaz Eftekhar, Alexander Sax, Jitendra Malik, and Amir Zamir.
\newblock Omnidata: A scalable pipeline for making multi-task mid-level vision
  datasets from 3d scans.
\newblock In {\em Proceedings of the IEEE/CVF Conference on Computer Vision and
  Pattern Recognition (CVPR)}, pages 10786--10796, 2021.

\bibitem{OmmerBjrn2020TamingTF}
Patrick Esser, Robin Rombach, and Bjorn Ommer.
\newblock Taming transformers for high-resolution image synthesis.
\newblock In {\em Proceedings of the IEEE/CVF Conference on Computer Vision and
  Pattern Recognition (CVPR)}, pages 12873--12883, 2021.

\bibitem{forte2020fbamatting}
Marco Forte and Fran{\c{c}}ois Piti{\'e}.
\newblock F, b, alpha matting.
\newblock {\em arXiv.org}, 2003.07711, 2020.

\bibitem{yu2021plenoxels}
Sara Fridovich-Keil, Alex Yu, Matthew Tancik, Qinhong Chen, Benjamin Recht, and
  Angjoo Kanazawa.
\newblock Plenoxels: Radiance fields without neural networks.
\newblock In {\em Proceedings of the IEEE/CVF Conference on Computer Vision and
  Pattern Recognition (CVPR)}, pages 5501--5510, 2022.

\bibitem{fu20213d}
Huan Fu, Rongfei Jia, Lin Gao, Mingming Gong, Binqiang Zhao, Steve Maybank, and
  Dacheng Tao.
\newblock 3d-future: 3d furniture shape with texture.
\newblock {\em International Journal of Computer Vision (IJCV)},
  129(12):3313--3337, 2021.

\bibitem{MatheusGadelha20163DSI}
Matheus Gadelha, Subhransu Maji, and Rui Wang.
\newblock 3d shape induction from 2d views of multiple objects.
\newblock In {\em Proceedings of the International Conference on 3D Vision
  (3DV)}, pages 402--411, 2017.

\bibitem{JunGao2022GET3DAG}
Jun Gao, Tianchang Shen, Zian Wang, Wenzheng Chen, Kangxue Yin, Daiqing Li, Or
  Litany, Zan Gojcic, and Sanja Fidler.
\newblock Get3d: A generative model of high quality 3d textured shapes learned
  from images.
\newblock {\em arXiv.org}, 2209.11163, 2022.

\bibitem{gao2022get3d}
Jun Gao, Tianchang Shen, Zian Wang, Wenzheng Chen, Kangxue Yin, Daiqing Li, Or
  Litany, Zan Gojcic, and Sanja Fidler.
\newblock Get3d: A generative model of high quality 3d textured shapes learned
  from images.
\newblock In {\em Advances in Neural Information Processing Systems (NIPS)},
  2022.

\bibitem{goyal2021simpleview}
Ankit Goyal, Hei Law, Bowei Liu, Alejandro Newell, and Jia Deng.
\newblock Revisiting point cloud shape classification with a simple and
  effective baseline.
\newblock In {\em Proceedings of the International Conference on Machine
  learning (ICML)}, pages 3809--3820, 2021.

\bibitem{JiataoGu2021StyleNeRFAS}
Jiatao Gu, Lingjie Liu, Peng Wang, and Christian Theobalt.
\newblock Stylenerf: A style-based 3d-aware generator for high-resolution image
  synthesis.
\newblock {\em Proceedings of the International Conference on Learning
  Representations (ICLR)}, 2022.

\bibitem{guo2020pct}
Meng-Hao Guo, Jun-Xiong Cai, Zheng-Ning Liu, Tai-Jiang Mu, Ralph~R Martin, and
  Shi-Min Hu.
\newblock Pct: Point cloud transformer.
\newblock {\em Computational Visual Media}, 7(2):187--199, 2021.

\bibitem{gupta2019lvis}
Agrim Gupta, Piotr Dollar, and Ross Girshick.
\newblock {LVIS}: A dataset for large vocabulary instance segmentation.
\newblock In {\em Proceedings of the IEEE/CVF Conference on Computer Vision and
  Pattern Recognition (CVPR)}, pages 5356--5364, 2019.

\bibitem{ZekunHao2021GANcraftU3}
Zekun Hao, Arun Mallya, Serge Belongie, and Ming-Yu Liu.
\newblock Gancraft: Unsupervised 3d neural rendering of minecraft worlds.
\newblock In {\em Proceedings of the IEEE/CVF International Conference on
  Computer Vision (ICCV)}, pages 14072--14082, 2021.

\bibitem{PhilippHenzler2018EscapingPC}
Philipp Henzler, Niloy~J Mitra, and Tobias Ritschel.
\newblock Escaping plato's cave: 3d shape from adversarial rendering.
\newblock In {\em Proceedings of the IEEE/CVF International Conference on
  Computer Vision (ICCV)}, pages 9984--9993, 2019.

\bibitem{henzler2021unsupervised}
Philipp Henzler, Jeremy Reizenstein, Patrick Labatut, Roman Shapovalov, Tobias
  Ritschel, Andrea Vedaldi, and David Novotny.
\newblock Unsupervised learning of 3d object categories from videos in the
  wild.
\newblock In {\em Proceedings of the IEEE/CVF Conference on Computer Vision and
  Pattern Recognition}, pages 4700--4709, 2021.

\bibitem{heusel2017gans}
Martin Heusel, Hubert Ramsauer, Thomas Unterthiner, Bernhard Nessler, and Sepp
  Hochreiter.
\newblock Gans trained by a two time-scale update rule converge to a local nash
  equilibrium.
\newblock In {\em Advances in Neural Information Processing Systems (NIPS)},
  volume~30, 2017.

\bibitem{XunHuang2022MultimodalCI}
Xun Huang, Arun Mallya, Ting-Chun Wang, and Ming-Yu Liu.
\newblock Multimodal conditional image synthesis with product-of-experts gans.
\newblock In {\em Proceedings of the European Conference on Computer Vision
  (ECCV)}, pages 91--109, 2022.

\bibitem{MoritzIbing2022OctreeTA}
Moritz Ibing, Gregor Kobsik, and Leif Kobbelt.
\newblock Octree transformer: Autoregressive 3d shape generation on
  hierarchically structured sequences.
\newblock {\em arXiv.org}, 2111.12480, 2022.

\bibitem{jiang2020sdfdiff}
Yue Jiang, Dantong Ji, Zhizhong Han, and Matthias Zwicker.
\newblock Sdfdiff: Differentiable rendering of signed distance fields for 3d
  shape optimization.
\newblock In {\em Proceedings of the IEEE/CVF Conference on Computer Vision and
  Pattern Recognition (CVPR)}, pages 1251--1261, 2020.

\bibitem{TeroKarras2021AliasFreeGA}
Tero Karras, Miika Aittala, Samuli Laine, Erik H{\"a}rk{\"o}nen, Janne
  Hellsten, Jaakko Lehtinen, and Timo Aila.
\newblock Alias-free generative adversarial networks.
\newblock In {\em Advances in Neural Information Processing Systems (NIPS)},
  volume~34, pages 852--863, 2021.

\bibitem{TeroKarras2018ASG}
Tero Karras, Samuli Laine, and Timo Aila.
\newblock A style-based generator architecture for generative adversarial
  networks.
\newblock In {\em Proceedings of the IEEE/CVF Conference on Computer Vision and
  Pattern Recognition (CVPR)}, pages 4401--4410, 2019.

\bibitem{LehtinenJaakko2019AnalyzingAI}
Tero Karras, Samuli Laine, Miika Aittala, Janne Hellsten, Jaakko Lehtinen, and
  Timo Aila.
\newblock Analyzing and improving the image quality of stylegan.
\newblock In {\em Proceedings of the IEEE/CVF Conference on Computer Vision and
  Pattern Recognition (CVPR)}, pages 8110--8119, 2020.

\bibitem{kellnhofer2021neural}
Petr Kellnhofer, Lars~C Jebe, Andrew Jones, Ryan Spicer, Kari Pulli, and Gordon
  Wetzstein.
\newblock Neural lumigraph rendering.
\newblock In {\em Proceedings of the IEEE/CVF Conference on Computer Vision and
  Pattern Recognition (CVPR)}, pages 4287--4297, 2021.

\bibitem{kim2021pointwolf}
Sihyeon Kim, Sanghyeok Lee, Dasol Hwang, Jaewon Lee, Seong~Jae Hwang, and
  Hyunwoo~J Kim.
\newblock Point cloud augmentation with weighted local transformations.
\newblock In {\em Proceedings of the IEEE/CVF International Conference on
  Computer Vision (ICCV)}, pages 548--557, 2021.

\bibitem{kingma2014adam}
Diederik~P Kingma and Jimmy Ba.
\newblock Adam: A method for stochastic optimization.
\newblock {\em arXiv.org}, 1412.6980, 2014.

\bibitem{kuznetsova2020open}
Alina Kuznetsova, Hassan Rom, Neil Alldrin, Jasper Uijlings, Ivan Krasin, Jordi
  Pont-Tuset, Shahab Kamali, Stefan Popov, Matteo Malloci, Alexander
  Kolesnikov, et~al.
\newblock The open images dataset v4.
\newblock {\em International Journal of Computer Vision (IJCV)},
  128(7):1956--1981, 2020.

\bibitem{lin2014microsoft}
Tsung-Yi Lin, Michael Maire, Serge Belongie, James Hays, Pietro Perona, Deva
  Ramanan, Piotr Doll{\'a}r, and C~Lawrence Zitnick.
\newblock Microsoft coco: Common objects in context.
\newblock In {\em Proceedings of the European Conference on Computer Vision
  (ECCV)}, pages 740--755, 2014.

\bibitem{liu2022akb48}
Liu Liu, Wenqiang Xu, Haoyuan Fu, Sucheng Qian, Qiaojun Yu, Yang Han, and Cewu
  Lu.
\newblock Akb-48: A real-world articulated object knowledge base.
\newblock In {\em Proceedings of the IEEE/CVF Conference on Computer Vision and
  Pattern Recognition (CVPR)}, pages 14809--14818, 2022.

\bibitem{liu2020dist}
Shaohui Liu, Yinda Zhang, Songyou Peng, Boxin Shi, Marc Pollefeys, and Zhaopeng
  Cui.
\newblock Dist: Rendering deep implicit signed distance function with
  differentiable sphere tracing.
\newblock In {\em Proceedings of the IEEE/CVF Conference on Computer Vision and
  Pattern Recognition (CVPR)}, pages 2019--2028, 2020.

\bibitem{liu2019rscnn}
Yongcheng Liu, Bin Fan, Shiming Xiang, and Chunhong Pan.
\newblock Relation-shape convolutional neural network for point cloud analysis.
\newblock In {\em Proceedings of the IEEE/CVF Conference on Computer Vision and
  Pattern Recognition (CVPR)}, pages 8895--8904, 2019.

\bibitem{YuanLiu2021NeuralRF}
Yuan Liu, Sida Peng, Lingjie Liu, Qianqian Wang, Peng Wang, Christian Theobalt,
  Xiaowei Zhou, and Wenping Wang.
\newblock Neural rays for occlusion-aware image-based rendering.
\newblock In {\em Proceedings of the IEEE/CVF Conference on Computer Vision and
  Pattern Recognition (CVPR)}, pages 7824--7833, 2022.

\bibitem{lombardi2019neural}
Stephen Lombardi, Tomas Simon, Jason Saragih, Gabriel Schwartz, Andreas
  Lehrmann, and Yaser Sheikh.
\newblock Neural volumes: Learning dynamic renderable volumes from images.
\newblock {\em arXiv.org}, 1906.07751, 2019.

\bibitem{XiaoxiaoLong2022SparseNeuSFG}
Xiaoxiao Long, Cheng Lin, Peng Wang, Taku Komura, and Wenping Wang.
\newblock Sparseneus: Fast generalizable neural surface reconstruction from
  sparse views.
\newblock {\em arXiv.org}, 2206.05737, 2022.

\bibitem{SebastianLunz2020InverseGG}
Sebastian Lunz, Yingzhen Li, Andrew Fitzgibbon, and Nate Kushman.
\newblock Inverse graphics gan: Learning to generate 3d shapes from
  unstructured 2d data.
\newblock {\em arXiv.org}, 2002.12674, 2020.

\bibitem{AndrewLuo2022SurfGenA3}
Andrew Luo, Tianqin Li, Wen-Hao Zhang, and Tai~Sing Lee.
\newblock Surfgen: Adversarial 3d shape synthesis with explicit surface
  discriminators.
\newblock In {\em Proceedings of the IEEE/CVF International Conference on
  Computer Vision (ICCV)}, pages 16238--16248, 2021.

\bibitem{LarsMescheder2018OccupancyNL}
Lars Mescheder, Michael Oechsle, Michael Niemeyer, Sebastian Nowozin, and
  Andreas Geiger.
\newblock Occupancy networks: Learning 3d reconstruction in function space.
\newblock In {\em Proceedings of the IEEE/CVF Conference on Computer Vision and
  Pattern Recognition (CVPR)}, pages 4460--4470, 2019.

\bibitem{mescheder2019occupancy}
Lars Mescheder, Michael Oechsle, Michael Niemeyer, Sebastian Nowozin, and
  Andreas Geiger.
\newblock Occupancy networks: Learning 3d reconstruction in function space.
\newblock In {\em Proceedings of the IEEE/CVF Conference on Computer Vision and
  Pattern Recognition (CVPR)}, pages 4460--4470, 2019.

\bibitem{BenMildenhall2021NeRFIT}
Ben Mildenhall, Peter Hedman, Ricardo Martin-Brualla, Pratul~P Srinivasan, and
  Jonathan~T Barron.
\newblock Nerf in the dark: High dynamic range view synthesis from noisy raw
  images.
\newblock In {\em Proceedings of the IEEE/CVF Conference on Computer Vision and
  Pattern Recognition (CVPR)}, pages 16190--16199, 2022.

\bibitem{mildenhall2020nerf}
Ben Mildenhall, Pratul~P Srinivasan, Matthew Tancik, Jonathan~T Barron, Ravi
  Ramamoorthi, and Ren Ng.
\newblock Nerf: Representing scenes as neural radiance fields for view
  synthesis.
\newblock In {\em Proceedings of the European Conference on Computer Vision
  (ECCV)}, pages 405--421, 2020.

\bibitem{KaichunMo2019StructureNetHG}
Kaichun Mo, Paul Guerrero, Li Yi, Hao Su, Peter Wonka, Niloy~J. Mitra, and
  Leonidas~J. Guibas.
\newblock Structurenet: Hierarchical graph networks for 3d shape generation.
\newblock {\em arXiv.org}, 1908.00575, 2019.

\bibitem{mueller2022instant}
Thomas M\"uller, Alex Evans, Christoph Schied, and Alexander Keller.
\newblock Instant neural graphics primitives with a multiresolution hash
  encoding.
\newblock {\em ACM Transactions on Graphics}, 2022.

\bibitem{CharlieNash2020PolyGenAA}
Charlie Nash, Yaroslav Ganin, SM~Ali Eslami, and Peter Battaglia.
\newblock Polygen: An autoregressive generative model of 3d meshes.
\newblock In {\em Proceedings of the International Conference on Machine
  learning (ICML)}, pages 7220--7229, 2020.

\bibitem{MichaelNiemeyer2020GIRAFFERS}
Michael Niemeyer and Andreas Geiger.
\newblock Giraffe: Representing scenes as compositional generative neural
  feature fields.
\newblock In {\em Proceedings of the IEEE/CVF Conference on Computer Vision and
  Pattern Recognition (CVPR)}, pages 11453--11464, 2021.

\bibitem{niemeyer2020differentiable}
Michael Niemeyer, Lars Mescheder, Michael Oechsle, and Andreas Geiger.
\newblock Differentiable volumetric rendering: Learning implicit 3d
  representations without 3d supervision.
\newblock In {\em Proceedings of the IEEE/CVF Conference on Computer Vision and
  Pattern Recognition (CVPR)}, pages 3504--3515, 2020.

\bibitem{oechsle2021unisurf}
Michael Oechsle, Songyou Peng, and Andreas Geiger.
\newblock Unisurf: Unifying neural implicit surfaces and radiance fields for
  multi-view reconstruction.
\newblock In {\em Proceedings of the IEEE/CVF International Conference on
  Computer Vision (ICCV)}, pages 5589--5599, 2021.

\bibitem{RoyOrEl2022StyleSDFH3}
Roy Or-El, Xuan Luo, Mengyi Shan, Eli Shechtman, Jeong~Joon Park, and Ira
  Kemelmacher-Shlizerman.
\newblock Stylesdf: High-resolution 3d-consistent image and geometry
  generation.
\newblock In {\em Proceedings of the IEEE/CVF Conference on Computer Vision and
  Pattern Recognition (CVPR)}, pages 13503--13513, 2022.

\bibitem{park2019deepsdf}
Jeong~Joon Park, Peter Florence, Julian Straub, Richard Newcombe, and Steven
  Lovegrove.
\newblock Deepsdf: Learning continuous signed distance functions for shape
  representation.
\newblock In {\em Proceedings of the IEEE/CVF Conference on Computer Vision and
  Pattern Recognition (CVPR)}, pages 165--174, 2019.

\bibitem{TaesungPark2019SemanticIS}
Taesung Park, Ming-Yu Liu, Ting-Chun Wang, and Jun-Yan Zhu.
\newblock Semantic image synthesis with spatially-adaptive normalization.
\newblock In {\em Proceedings of the IEEE/CVF Conference on Computer Vision and
  Pattern Recognition (CVPR)}, pages 2337--2346, 2019.

\bibitem{DarioPavllo2021LearningGM}
Dario Pavllo, Jonas Kohler, Thomas Hofmann, and Aurelien Lucchi.
\newblock Learning generative models of textured 3d meshes from real-world
  images.
\newblock In {\em Proceedings of the IEEE/CVF International Conference on
  Computer Vision (ICCV)}, pages 13879--13889, 2021.

\bibitem{qi2016pointnet}
Charles~R Qi, Hao Su, Kaichun Mo, and Leonidas~J Guibas.
\newblock Pointnet: deep learning on point sets for 3d classification and
  segmentation. corr abs/1612.00593 (2016).
\newblock In {\em Proceedings of the IEEE/CVF Conference on Computer Vision and
  Pattern Recognition (CVPR)}, pages 652--660, 2017.

\bibitem{qi2017pointnetplusplus}
Charles~Ruizhongtai Qi, Li Yi, Hao Su, and Leonidas~J Guibas.
\newblock Pointnet++: Deep hierarchical feature learning on point sets in a
  metric space.
\newblock In {\em Advances in Neural Information Processing Systems (NIPS)},
  volume~30, 2017.

\bibitem{Qin_2020_PR}
Xuebin Qin, Zichen Zhang, Chenyang Huang, Masood Dehghan, Osmar~R Zaiane, and
  Martin Jagersand.
\newblock U2-net: Going deeper with nested u-structure for salient object
  detection.
\newblock {\em Pattern Recognition}, 106:107404, 2020.

\bibitem{reizenstein2021co3d}
Jeremy Reizenstein, Roman Shapovalov, Philipp Henzler, Luca Sbordone, Patrick
  Labatut, and David Novotny.
\newblock Common objects in 3d: Large-scale learning and evaluation of
  real-life 3d category reconstruction.
\newblock In {\em Proceedings of the IEEE/CVF International Conference on
  Computer Vision (ICCV)}, pages 10901--10911, 2021.

\bibitem{ren2022modelnet-c}
Jiawei Ren, Liang Pan, and Ziwei Liu.
\newblock Benchmarking and analyzing point cloud classification under
  corruptions.
\newblock In {\em Proceedings of the International Conference on Machine
  learning (ICML)}, 2022.

\bibitem{saito2019pifu}
Shunsuke Saito, Zeng Huang, Ryota Natsume, Shigeo Morishima, Angjoo Kanazawa,
  and Hao Li.
\newblock Pifu: Pixel-aligned implicit function for high-resolution clothed
  human digitization.
\newblock In {\em Proceedings of the IEEE/CVF International Conference on
  Computer Vision (ICCV)}, pages 2304--2314, 2019.

\bibitem{schonberger2016sfm}
Johannes~L Schonberger and Jan-Michael Frahm.
\newblock Structure-from-motion revisited.
\newblock In {\em Proceedings of the IEEE/CVF Conference on Computer Vision and
  Pattern Recognition (CVPR)}, pages 4104--4113, 2016.

\bibitem{KatjaSchwarz2020GRAFGR}
Katja Schwarz, Yiyi Liao, Michael Niemeyer, and Andreas Geiger.
\newblock Graf: Generative radiance fields for 3d-aware image synthesis.
\newblock In {\em Advances in Neural Information Processing Systems (NIPS)},
  volume~33, pages 20154--20166, 2020.

\bibitem{KatjaSchwarz2022VoxGRAFF3}
Katja Schwarz, Axel Sauer, Michael Niemeyer, Yiyi Liao, and Andreas Geiger.
\newblock Voxgraf: Fast 3d-aware image synthesis with sparse voxel grids.
\newblock {\em arXiv.org}, 2206.07695, 2022.

\bibitem{shao2019objects365}
Shuai Shao, Zeming Li, Tianyuan Zhang, Chao Peng, Gang Yu, Xiangyu Zhang, Jing
  Li, and Jian Sun.
\newblock Objects365: A large-scale, high-quality dataset for object detection.
\newblock In {\em Proceedings of the IEEE/CVF International Conference on
  Computer Vision (ICCV)}, pages 8430--8439, 2019.

\bibitem{sitzmann2019srns}
Vincent Sitzmann, Michael Zollh{\"o}fer, and Gordon Wetzstein.
\newblock Scene representation networks: Continuous 3d-structure-aware neural
  scene representations.
\newblock In {\em Advances in Neural Information Processing Systems (NIPS)},
  volume~32, 2019.

\bibitem{EdwardJSmith2017ImprovedAS}
Edward~J Smith and David Meger.
\newblock Improved adversarial systems for 3d object generation and
  reconstruction.
\newblock In {\em Proceedings of the Conference on Robot Learning (CoRL)},
  pages 87--96, 2017.

\bibitem{stojanov2021toys4k}
Stefan Stojanov, Anh Thai, and James~M Rehg.
\newblock Using shape to categorize: Low-shot learning with an explicit shape
  bias.
\newblock In {\em Proceedings of the IEEE/CVF Conference on Computer Vision and
  Pattern Recognition (CVPR)}, pages 1798--1808, 2021.

\bibitem{sun2021direct}
Cheng Sun, Min Sun, and Hwann-Tzong Chen.
\newblock Direct voxel grid optimization: Super-fast convergence for radiance
  fields reconstruction.
\newblock In {\em Proceedings of the IEEE/CVF Conference on Computer Vision and
  Pattern Recognition (CVPR)}, pages 5459--5469, 2022.

\bibitem{taghanaki2020robustpointset}
Saeid~Asgari Taghanaki, Jieliang Luo, Ran Zhang, Ye Wang, Pradeep~Kumar
  Jayaraman, and Krishna~Murthy Jatavallabhula.
\newblock Robustpointset: A dataset for benchmarking robustness of point cloud
  classifiers.
\newblock {\em arXiv.org}, 2011.11572, 2020.

\bibitem{toussaint2022hal}
Briac Toussaint, Maxime Genisson, and Jean-S{\'e}bastien Franco.
\newblock {Fast Gradient Descent for Surface Capture Via Differentiable
  Rendering}.
\newblock In {\em Proceedings of the International Conference on 3D Vision
  (3DV)}, pages 1--10, 2022.

\bibitem{uy2019revisiting}
Mikaela~Angelina Uy, Quang-Hieu Pham, Binh-Son Hua, Thanh Nguyen, and Sai-Kit
  Yeung.
\newblock Revisiting point cloud classification: A new benchmark dataset and
  classification model on real-world data.
\newblock In {\em Proceedings of the IEEE/CVF International Conference on
  Computer Vision (ICCV)}, pages 1588--1597, 2019.

\bibitem{MukundVarma2022IsAA}
Mukund Varma, Peihao Wang, Xuxi Chen, Tianlong Chen, Subhashini Venugopalan,
  Zhangyang Wang, and Madras.
\newblock Is attention all nerf needs?
\newblock {\em arXiv.org}, 2207.13298, 2022.

\bibitem{DorVerbin2022RefNeRFSV}
Dor Verbin, Peter Hedman, Ben Mildenhall, Todd Zickler, Jonathan~T Barron, and
  Pratul~P Srinivasan.
\newblock Ref-nerf: Structured view-dependent appearance for neural radiance
  fields.
\newblock In {\em Proceedings of the IEEE/CVF Conference on Computer Vision and
  Pattern Recognition (CVPR)}, pages 5481--5490, 2022.

\bibitem{wang2021neus}
Peng Wang, Lingjie Liu, Yuan Liu, Christian Theobalt, Taku Komura, and Wenping
  Wang.
\newblock Neus: Learning neural implicit surfaces by volume rendering for
  multi-view reconstruction.
\newblock In {\em Advances in Neural Information Processing Systems (NIPS)},
  volume~34, pages 27171--27183, 2021.

\bibitem{QianqianWang2021IBRNetLM}
Qianqian Wang, Zhicheng Wang, Kyle Genova, Pratul~P Srinivasan, Howard Zhou,
  Jonathan~T Barron, Ricardo Martin-Brualla, Noah Snavely, and Thomas
  Funkhouser.
\newblock Ibrnet: Learning multi-view image-based rendering.
\newblock In {\em Proceedings of the IEEE/CVF Conference on Computer Vision and
  Pattern Recognition (CVPR)}, pages 4690--4699, 2021.

\bibitem{wang2019dgcnn}
Yue Wang, Yongbin Sun, Ziwei Liu, Sanjay~E Sarma, Michael~M Bronstein, and
  Justin~M Solomon.
\newblock Dynamic graph cnn for learning on point clouds.
\newblock {\em ACM Transactions on Graphics}, 38(5):1--12, 2019.

\bibitem{wang2004image}
Zhou Wang, Alan~C Bovik, Hamid~R Sheikh, and Eero~P Simoncelli.
\newblock Image quality assessment: from error visibility to structural
  similarity.
\newblock {\em IEEE Transactions on Image Processing (TIP)}, 13(4):600--612,
  2004.

\bibitem{JiajunWu2016LearningAP}
Jiajun Wu, Chengkai Zhang, Tianfan Xue, Bill Freeman, and Josh Tenenbaum.
\newblock Learning a probabilistic latent space of object shapes via 3d
  generative-adversarial modeling.
\newblock In {\em Advances in Neural Information Processing Systems (NIPS)},
  volume~29, 2016.

\bibitem{wu2022voxurf}
Tong Wu, Jiaqi Wang, Xingang Pan, Xudong Xu, Christian Theobalt, Ziwei Liu, and
  Dahua Lin.
\newblock Voxurf: Voxel-based efficient and accurate neural surface
  reconstruction.
\newblock {\em arXiv.org}, 2208.12697, 2022.

\bibitem{wu20153d}
Zhirong Wu, Shuran Song, Aditya Khosla, Fisher Yu, Linguang Zhang, Xiaoou Tang,
  and Jianxiong Xiao.
\newblock 3d shapenets: A deep representation for volumetric shapes.
\newblock In {\em Proceedings of the IEEE/CVF Conference on Computer Vision and
  Pattern Recognition (CVPR)}, pages 1912--1920, 2015.

\bibitem{xiang2021curvenet}
Tiange Xiang, Chaoyi Zhang, Yang Song, Jianhui Yu, and Weidong Cai.
\newblock Walk in the cloud: Learning curves for point clouds shape analysis.
\newblock In {\em Proceedings of the IEEE/CVF International Conference on
  Computer Vision (ICCV)}, pages 915--924, 2021.

\bibitem{xu2021paconv}
Mutian Xu, Runyu Ding, Hengshuang Zhao, and Xiaojuan Qi.
\newblock Paconv: Position adaptive convolution with dynamic kernel assembling
  on point clouds.
\newblock In {\em Proceedings of the IEEE/CVF Conference on Computer Vision and
  Pattern Recognition (CVPR)}, pages 3173--3182, 2021.

\bibitem{xu2021gdanet}
Mutian Xu, Junhao Zhang, Zhipeng Zhou, Mingye Xu, Xiaojuan Qi, and Yu Qiao.
\newblock Learning geometry-disentangled representation for complementary
  understanding of 3d object point cloud.
\newblock In {\em Proceedings of the Conference on Artificial Intelligence
  (AAAI)}, volume~35, pages 3056--3064, 2021.

\bibitem{YinghaoXu20223DawareIS}
Yinghao Xu, Sida Peng, Ceyuan Yang, Yujun Shen, and Bolei Zhou.
\newblock 3d-aware image synthesis via learning structural and textural
  representations.
\newblock In {\em Proceedings of the IEEE/CVF Conference on Computer Vision and
  Pattern Recognition (CVPR)}, pages 18430--18439, 2022.

\bibitem{GuandaoYang2019PointFlow3P}
Guandao Yang, Xun Huang, Zekun Hao, Ming-Yu Liu, Serge Belongie, and Bharath
  Hariharan.
\newblock Pointflow: 3d point cloud generation with continuous normalizing
  flows.
\newblock In {\em Proceedings of the IEEE/CVF International Conference on
  Computer Vision (ICCV)}, pages 4541--4550, 2019.

\bibitem{yao2020blendedmvs}
Yao Yao, Zixin Luo, Shiwei Li, Jingyang Zhang, Yufan Ren, Lei Zhou, Tian Fang,
  and Long Quan.
\newblock Blendedmvs: A large-scale dataset for generalized multi-view stereo
  networks.
\newblock In {\em Proceedings of the IEEE/CVF Conference on Computer Vision and
  Pattern Recognition (CVPR)}, pages 1790--1799, 2020.

\bibitem{yariv2021volume}
Lior Yariv, Jiatao Gu, Yoni Kasten, and Yaron Lipman.
\newblock Volume rendering of neural implicit surfaces.
\newblock In {\em Advances in Neural Information Processing Systems (NIPS)},
  volume~34, pages 4805--4815, 2021.

\bibitem{yariv2020multiview}
Lior Yariv, Yoni Kasten, Dror Moran, Meirav Galun, Matan Atzmon, Basri Ronen,
  and Yaron Lipman.
\newblock Multiview neural surface reconstruction by disentangling geometry and
  appearance.
\newblock In {\em Advances in Neural Information Processing Systems (NIPS)},
  volume~33, pages 2492--2502, 2020.

\bibitem{AlexYu2021pixelNeRFNR}
Alex Yu, Vickie Ye, Matthew Tancik, and Angjoo Kanazawa.
\newblock pixelnerf: Neural radiance fields from one or few images.
\newblock In {\em Proceedings of the IEEE/CVF Conference on Computer Vision and
  Pattern Recognition (CVPR)}, pages 4578--4587, 2021.

\bibitem{ZehaoYu2022MonoSDFEM}
Zehao Yu, Songyou Peng, Michael Niemeyer, Torsten Sattler, and Andreas Geiger.
\newblock Monosdf: Exploring monocular geometric cues for neural implicit
  surface reconstruction.
\newblock In {\em Advances in Neural Information Processing Systems (NIPS)},
  2022.

\bibitem{zhang2021learning}
Jingyang Zhang, Yao Yao, and Long Quan.
\newblock Learning signed distance field for multi-view surface reconstruction.
\newblock In {\em Proceedings of the IEEE/CVF International Conference on
  Computer Vision (ICCV)}, pages 6525--6534, 2021.

\bibitem{zhang2018unreasonable}
Richard Zhang, Phillip Isola, Alexei~A Efros, Eli Shechtman, and Oliver Wang.
\newblock The unreasonable effectiveness of deep features as a perceptual
  metric.
\newblock In {\em Proceedings of the IEEE/CVF Conference on Computer Vision and
  Pattern Recognition (CVPR)}, pages 586--595, 2018.

\bibitem{LinqiZhou20213DSG}
Linqi Zhou, Yilun Du, and Jiajun Wu.
\newblock 3d shape generation and completion through point-voxel diffusion.
\newblock In {\em Proceedings of the IEEE/CVF International Conference on
  Computer Vision (ICCV)}, pages 5826--5835, 2021.

\bibitem{PengZhou2021CIPS3DA3}
Peng Zhou, Lingxi Xie, Bingbing Ni, and Qi Tian.
\newblock Cips-3d: A 3d-aware generator of gans based on
  conditionally-independent pixel synthesis.
\newblock {\em arXiv.org}, 2110.09788, 2021.

\bibitem{zhou2018open3d}
Qian-Yi Zhou, Jaesik Park, and Vladlen Koltun.
\newblock {Open3D}: {A} modern library for {3D} data processing.
\newblock {\em arXiv.org}, 1801.09847, 2018.

\end{thebibliography}
}

\end{document}